\documentclass[runningheads]{llncs}

\usepackage{eccv}

\pdfoutput=1
\usepackage{eccvabbrv}
\usepackage{multirow}
\usepackage{graphicx}
\usepackage{booktabs}
\usepackage{colortbl}
\usepackage{pifont}
\usepackage{caption}
\usepackage{wrapfig}
\usepackage{subcaption}

\usepackage{floatrow}
\newfloatcommand{capbtabbox}{table}[][\FBwidth]

\usepackage[accsupp]{axessibility}  

\usepackage[pagebackref,breaklinks,colorlinks,citecolor=eccvblue]{hyperref}

\usepackage{orcidlink}

\begin{document}

\title{Facial Affective Behavior Analysis with Instruction Tuning} 

\titlerunning{Facial Affective Behavior Analysis with Instruction Tuning}

\author{Yifan Li\inst{1} \and
Anh Dao\inst{1} \and
Wentao Bao\inst{1} \and
Zhen Tan\inst{2} \and
Tianlong Chen\inst{3,4,5} \and
Huan Liu\inst{2} \and
Yu Kong\inst{1}
}

\authorrunning{Y. Li  et al.}

\institute{Michigan State University,
\email{\{liyifa11, anhdao, baowenta, yukong\}@msu.edu}\\
 \and
Arizona State University,
\email{\{ztan36, huanliu\}@asu.edu}\\  
 \and
University of North Carolina at Chapel Hill\\
 \and
Massachusetts Institute of Technology,
\email{tianlong@mit.edu}\\
 \and
Harvard University
}

\maketitle

\begin{abstract}
    Facial affective behavior analysis (FABA) is crucial for understanding human mental states from images. 
    However, traditional approaches primarily deploy models to discriminate among discrete emotion categories, 
    and lack the fine granularity and reasoning capability for complex facial behaviors.
    The advent of Multi-modal Large Language Models (MLLMs) has been proven successful in general visual understanding tasks. However, directly harnessing MLLMs for FABA is challenging due to the scarcity of datasets and benchmarks, neglecting facial prior knowledge, and low training efficiency. 
    To address these challenges, we introduce (\textit{\textbf{i}}) an instruction-following dataset for two FABA tasks, \ie, facial emotion and action unit recognition, (\textit{\textbf{ii}})
     a benchmark \textit{FABA-Bench} with a new metric considering both recognition and generation ability, and (\textit{\textbf{iii}}) a new MLLM \textit{EmoLA} as a strong baseline to the community.
    Our initiative on the dataset and benchmarks reveal the nature and rationale of 
    facial affective behaviors, \ie, fine-grained facial movement, interpretability, and reasoning.  
    Moreover, to build an effective and efficient FABA MLLM, 
    we introduce a facial prior expert module with face structure knowledge and a low-rank adaptation module into
    pre-trained MLLM. 
    We conduct extensive experiments on FABA-Bench and four commonly-used FABA datasets.
    The results demonstrate that the proposed facial prior expert can boost the performance and EmoLA achieves the best results on our FABA-Bench.  On commonly-used FABA datasets, EmoLA is competitive rivaling task-specific state-of-the-art models. The dataset and codes are available: \url{https://johnx69.github.io/FABA/}.
    
    
    \keywords{Facial affective behavior analysis \and Emotion recognition \and AU recognition \and Multi-modal large language model }
\end{abstract}
\section{Introduction}

Facial affective behavior analysis (FABA) \cite{kollias2020analysing,martinez2017automatic} such as facial emotion recognition (FER) \cite{li2020deep,lopes2017facial}, and action unit recognition (AUR) \cite{ekman1978facial}, aims to recognize 
facial expressions and movements, which are critical to understanding an individual's emotional states and intentions~\cite{grandjean2008conscious}. 
FABA has emerged as a burgeoning field with potential across multiple domains. For example, in psychology, FABA can aid therapists by offering real-time insights into a patient's unspoken emotions through facial expression analysis, thereby enhancing therapeutic outcomes~\cite{pixton2011happy}. In education, it improves e-learning experiences by adjusting content delivery based on students' facial cues, indicating engagement or confusion~\cite{saneiro2014towards}.
\begin{figure}[t]
    \centering
    \includegraphics[width=0.98\textwidth]{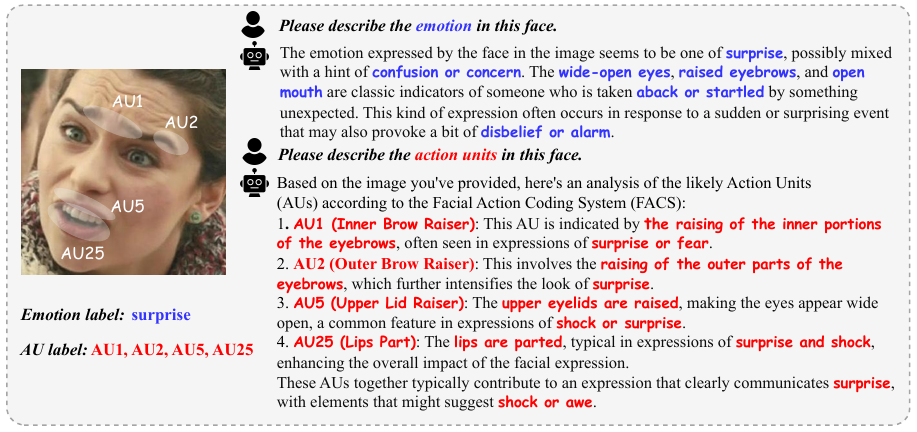}
    \caption{An illustration of FABA-Instruct annotations. FABA-Instruct can provide fine-grained emotion and AU descriptions, which not only include the reasoning process about the facial movements but also present the inference to the emotion. Furthermore, compared to traditional category labels, FABA-Instruct has more abundant expressions to describe complex, nuanced, exaggerated, and undefined affective behaviors. }
    \label{fig:intro}
\end{figure}

Despite promising progress being made, most of the existing FABA approaches \cite{wang2020region,li2018occlusion,shao2021jaa,niu2019local} are based on discriminative models, which treat FER or AUR as a multi-class or multi-label classification task. Such approaches tend to induce shortcomings like coarse-grained emotional descriptions, inability to describe complex emotions, and lack of reasoning ability. Those limitations hinder the applications of FABA, for instance, in providing nuanced feedback to therapists about a patient's emotional nuances in psychology, or in accurately adapting educational content based on subtle student reactions in e-learning environments. To counteract these drawbacks, we are motivated by the success of recent multi-modal large language models (MLLMs)~\cite{zhang2024mm,wang2024exploring}, because of their evidenced ability to describe and reason over fine-grained and complex visual cues by instruction tuning after large-scale pre-training~\cite{liu2024visual}. 
In practice, MLLMs transform the discriminative task into a sequence-to-sequence generative one \cite{chen2021pix2seq,cho2021unifying} based on large language models (LLMs) \cite{touvron2023llama,jiang2023mistral,touvron2023llama2}. MLLMs have shown great capability on various visual understanding tasks such as visual question answering~\cite{liu2024visual,liu2023improved,zhu2023minigpt}, captioning \cite{li2022blip,li2023blip}, grounding \cite{you2023ferret,chen2023shikra}, segmentation \cite{lai2023lisa}, \etc. 

However, there exist three major challenges in FABA tasks when deploying MLLMs. (1) There is no suitable FABA dataset for MLLMs to perform instruction tuning. Existing FABA datasets have either coarse-grained annotations \cite{kollias2018aff,li2017reliable,li2019reliable,zhang2013high,zhang2014bp4d,mavadati2013disfa,fabian2016emotionet} or limited emotion descriptions \cite{liu2022mafw} for video clips, since it's labor-intensive and expensive to manually annotate large-scale fine-grained FABA descriptions especially for AUR task. (2) There is an increasing number of MLLMs, 
how to select suitable MLLMs for FABA remains unknown. Existing metrics for evaluating MLLMs are language-oriented, without specific consideration of language usage in FABA tasks. 
(3) Furthermore, image features from vision encoders like CLIP \cite{radford2021learning} of current MLLMs struggle to 
capture facial structure information such as facial landmarks, leaving the impact of facial priors on FABA tasks unexplored. 
Fine-tuning the entire model hosting billions of parameters for these features leads to prohibitive computational costs.


To solve these challenges, we propose an instruction-following FABA dataset ``{FABA-Instruct}''. It includes 19$K$ in-the-wild aligned face images with 30$K$ fine-grained emotion and AU annotations using GPT4V enabling instruction tuning (\cref{fig:intro}). Based on this dataset, we propose a new benchmark ``{FABA-Bench}'' for evaluating both the visual recognition and text generation performance of various MLLMs on FABA tasks. Moreover, we introduce an efficient MLLM ``{EmoLA}'' for FABA tasks by incorporating a low-rank adaptation method and a facial prior expert to a pre-trained MLLM like LLaVA-1.5 \cite{liu2023improved}. Specifically, to obtain the facial prior knowledge, we utilize a pre-trained face alignment encoder to extract the facial landmark features, which are complementary to 
the vision encoder. To mitigate the computational cost, the LoRA \cite{hu2021lora} method is adopted in training such that only the parameter residual is learned through low-rank matrices. In this paper, we take the earliest trial of the FABA tasks with instruction tuning over MLLMs, which shed light on 
both FABA and MLLMs research community. Our contributions are summarized in three-folds:

\begin{itemize}
    \item \textit{\textbf{Instruction-following FABA data}}. To our best knowledge, this is the first FABA dataset that enables instruction tuning. It reveals new aspects of FABA research topics and it will continuously bring the benefits of MLLMs to the FABA community. 
    \item \textit{\textbf{Instruction-following FABA benchmark}}. To evaluate the recognition and reasoning ability of different models on instruction-following FABA tasks, we introduce the {FABA-Bench} benchmark with a unique metric, \ie, REGE, allowing for both recognition and generation capability.

    \item \textbf{\textit{MLLM-based FABA architecture}}. To efficiently train on FABA tasks and utilize the facial prior knowledge, we introduce the {EmoLA} model which involves tuning LoRA on a pre-trained MLLM and incorporating a facial prior expert. We demonstrate the effectiveness of EmoLA on FABA-Instruct and four traditional FABA datasets. The results show that EmoLA achieves the best performance on FABA-Instruct and SOTA-comparable or even better results on the traditional FABA datasets. 
\end{itemize}

\section{Related work}

\subsection{Facial affective behavior analysis}
\textbf{Psychology perspective.} According to psychology research \cite{grandjean2008conscious}, two mechanisms can be utilized to model facial affective behaviors, \ie, emotion categories, and dimensional theory. According to Izard \cite{izard2013human} and Ekman \cite{ekman1999basic}, the basic emotions can be categorized into one of the several prototypes. Ekman \cite{ekman1978facial} also proposed to decompose the macroscopic affective behaviors into fine-grained Action Units (AUs) from the anatomical perspective. However, such discrete emotion representations may be not sufficient to capture complex and fine-grained emotions. Dimensional theory describes the continuous emotions from the Euclidean space perspective. Russel adopted two dimensions: arousal and valence \cite{russell1980circumplex} to represent pleasantness and degree of feelings. Although the dimensional theory allows for a more nuanced understanding of emotions, it is challenging to measure and recognize for humans. We argue that human-generated descriptions offer a superior way of characterizing facial affective behaviors (see \cref{fig:emo_analysis} and \cref{fig:au_analysis}), which not only capture the complexity and subtlety of affective behaviors but are also more accessible and quantifiable for humans.

\noindent \textbf{Methodology perspective.}
To better recognize the affective behaviors, current research mainly focuses on deep-learning-based techniques. These approaches can be categorized into three streams according to the task types, \ie, facial emotion recognition \cite{sun2020dynamic,wang2021oaenet,zhang2023learning,li2023decoupled,chen2023multivariate,zhang2023weakly}, action unit recognition \cite{tang2021piap,song2021dynamic,jacob2021facial,yang2021exploiting,chang2022knowledge,li2023recot,cui2023biomechanics,yin2024fg,zhang2024multimodal} and valance-arousal regression \cite{mollahosseini2017affectnet,zhao2020emotion,li2022affective,savchenko2022frame}. Existing methods focus on capturing the fine-grained facial movement via attention mechanism \cite{xie2019deep,wang2020region,wang2021oaenet,zhang2022learn,wen2023distract,xue2022vision}, improving generalization ability by introducing auxiliary information like facial landmarks \cite{niu2019local,tang2021piap} or extra data \cite{niu2019multi,li2023recot}, exploring the relationship among emotions or AUs \cite{li2019semantic,shao2019facial,zhao2018learning}, exploiting pre-trained model like self-supervised learning \cite{li2022affective,li2019self,xie2020self,sun2021emotion,chang2022knowledge}, and probing semantic information of affective behaviors \cite{yang2021exploiting} using CLIP \cite{radford2021learning,ViT}. However, these methods are mainly discriminative-based, which fail to generate fine-grained descriptions. By contrast, our method can generate detailed descriptions based on the prior knowledge from MLLM and the facial prior expert. 
\subsection{Multi-modal LLMs and efficient LLM adaptation} 
\textbf{Multi-modal LLMs.} Multi-modal LLMs are getting popular in multi-modal content understanding~\cite{chen2023shikra,wu2023nextgpt,dong2023dreamllm,wang2023cogvlm,lin2023vila}.
They are built on top of LLMs~\cite{flant5,touvron2023llama,vicuna2023,zhang2022opt,bai2023qwen,jiang2023mistral}, and transform visual (videos and images) and text data into a sequence of \emph{tokens} as input, resulting in generative modeling of downstream multi-modal understanding tasks by next token prediction. Specific to the image-based MLLMs, image tokens are typically encoded by CLIP vision transformer~\cite{radford2021learning,ViT}. Then, one of the major challenges in MLLM is how to project image token features into the language domain to better utilize the instruction-following capability of LLMs. In literature, Flamingo~\cite{alayrac2022flamingo} is an early work that uses cross-attention to build the interaction between images and text tokens. This design is followed by recent MLLMs~\cite{li2023otter,Emu,bai2023qwenvl,chen2023internvl}. Based on the cross-attention mechanism, Q-Former~\cite{li2023blip} was proved to be a superior visual-text projector and inspires recent line of MLLM research~\cite{zhu2023minigpt,chen2023xllm,instructblip,li2024up}. Recently, in contrast to the attention-based projector, LLaVA series~\cite{liu2024visual,liu2023improved} propose to use a simple MLP as the projector and resort to the LLMs to handle the visual-text interaction. 
In this paper, based on LLaVA-1.5~\cite{liu2023improved}, we developed an MLLM for affective behavior analysis and benchmarked multiple MLLM baselines. 

\noindent\textbf{Efficient LLM adaptation.} Though most existing multimodal LLMs show impressive performance on general visual-language tasks, how to adapt them to downstream applications is still challenging~\cite{xu2023parameter}, especially with limited instructional annotation data and computing resources. This raises the surge of research on parameter-efficient fine-tuning (PEFT)~\cite{houlsby2019parameter}, which avoids fine-tuning the entire LLM in training. According to~\cite{xu2023parameter}, existing PEFT methods can be categorized into addictive~\cite{houlsby2019parameter,lin2020exploring,he2021towards,ruckle2020adapterdrop,pfeiffer2020adapterfusion,mahabadi2021parameter,lester2021power,li2021prefix,liu2022few}, partial~\cite{zaken2021bitfit,guo2020parameter,xu2021raise,sung2021training}, reparametrized~\cite{hu2021lora,dettmers2024qlora,zhang2023adaptive,liu2023moelora}, hybrid~\cite{he2021towards,karimi2021compacter,mao2021unipelt,chen2023parameter}, and unified fine-tuning~\cite{wang2022adamix,he2022sparseadapter,yu2022unified,shen2021umec}. 
In this paper, we are interested in the re-parameterized fine-tuning as it gains increasing research attention. In this line, LoRA~\cite{hu2021lora} is a pioneering work, which is inspired by the fact that modal weight adaptation has a low \emph{intrinsic rank}. In training, LoRA keeps most LLM parameters frozen while only optimizing the low-rank factorized matrices of dense layers' residual. In our work, we empirically found the effectiveness of LoRA on the adaptation of MLLMs toward affective behavior analysis tasks.

\vspace{-10pt}
\section{Dataset and Benchmark}


\setlength{\tabcolsep}{1mm}
{ 
    \begin{figure}[t]        
    \begin{minipage}[t]{0.55\textwidth}
        \begin{table}[H]
        \renewcommand\arraystretch{1}
        \centering
        \caption{Existing FABA datasets.} \label{dataset_comparison} \centering
        \scalebox{0.63}{
        \begin{tabular}{lcccc}
          \toprule
          \textbf{Datasets}   &   \textbf{AU} & \textbf{Emotion} & \textbf{Annotation} & \textbf{In-the wild}  \\
          \midrule
          RAF-DB \cite{li2017reliable}& \ding{56} & \ding{51} & category & \ding{56} \\
          CK+ \cite{lucey2010extended}& \ding{56} & \ding{51} & category & \ding{56} \\
          MMI \cite{pantic2005web}& \ding{56} & \ding{51} & category & \ding{56} \\
          SFEW \cite{jiang2020dfew}& \ding{56} & \ding{51} & category & \ding{51} \\
          AffectNet \cite{mollahosseini2017affectnet} & \ding{56} & \ding{51} & category & \ding{51} \\
          MAFW \cite{liu2022mafw} & \ding{56} & \ding{51} & category \& short text & \ding{51} \\
          DFEW \cite{jiang2020dfew} & \ding{56} & \ding{51} & category & \ding{51} \\
          FERV39K \cite{wang2022ferv39k}  & \ding{56} & \ding{51} & category & \ding{51} \\
          FER2013 \cite{goodfellow2013challenges} & \ding{56} & \ding{51} & category & \ding{51} \\
          DISFA \cite{mavadati2013disfa} & \ding{51} & \ding{56} & category & \ding{56} \\
          GFT \cite{girard2017sayette} & \ding{51} & \ding{56} & category & \ding{56} \\
          CASME-II \cite{yan2014casme}   & \ding{51} & \ding{51} & category & \ding{56} \\
          BP4D \cite{zhang2014bp4d} & \ding{51} & \ding{51} & category & \ding{56} \\
          EmotioNet \cite{fabian2016emotionet} & \ding{51} & \ding{51} & category & \ding{51} \\
          AffWild2 \cite{kollias2018aff} & \ding{51} & \ding{51} & category & \ding{51} \\
          \rowcolor{gray!30} FABA-Instruct  & \ding{51} & \ding{51} & instruction \& description & \ding{51} \\
          \bottomrule 
       \end{tabular}}
       \end{table}
   \end{minipage}
   \hfill
    \begin{minipage}[t]{0.42\textwidth}
    \begin{table}[H]
        \caption{ FABA-Instruct statistics.} \label{tab:stat}\centering 
        \scalebox{0.69}{
        \begin{tabular}{lc}
          \toprule
          \textbf{Statistics}   &   \textbf{Value}  \\
          \midrule
          Total images & 19877 \\
          Emotion training samples & 19474 \\
          Emotion testing samples & 403 \\
          Emotion description average length & 50.47 \\
          AU training samples & 15838 \\
          AU testing samples & 325 \\
          AU description average length & 207.35 \\
          \bottomrule 
       \end{tabular}}
   \end{table}
    \vspace{-15mm}
    
   \begin{figure}[H]
       \centering
       \includegraphics[width=0.9\textwidth]{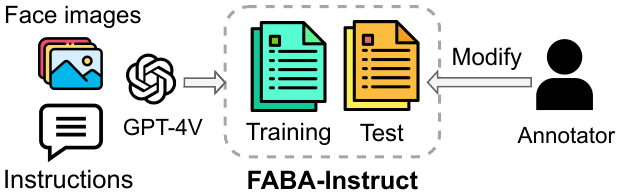}
       \vspace{-3mm}
       \caption{ \footnotesize FABA-Instruct annotation.}
       \label{fig:annotation_faba}
   \end{figure}
   \end{minipage}
    \end{figure} 
}

\subsection{Instruction-following FABA dataset}
In this section, we will present our instruction-following FABA dataset \textbf{FABA-Instruct} with two different FABA tasks, \ie, emotion recognition and AU recognition. Our dataset is different from existing FABA datasets \wrt tasks, annotation types, and the variety of images, which is presented in \cref{dataset_comparison}. This table illustrates that most of previous FABA datasets are either task-specific \cite{lucey2010extended, li2017reliable, pantic2005web, jiang2020dfew, mollahosseini2017affectnet, liu2022mafw, wang2022ferv39k, goodfellow2013challenges, zhang2014bp4d, mavadati2013disfa, girard2017sayette}, or laboratory collected \cite{lucey2010extended, pantic2005web, mavadati2013disfa, girard2017sayette, zhang2014bp4d}. More importantly, none of these FABA datasets have instruction-following annotations.
FABA-Instruct is the inaugural dataset to offer instruction-following annotations of both AU and emotion, specifically for in-the-wild face images. 

\noindent\textbf{Data construction.} The overall annotation pipeline is shown in \cref{fig:annotation_faba}. Specifically, we use 100 carefully designed templates as the instructions for querying GPT-4V \cite{lu2024gpt,tan2024large} on the emotions and AUs in the face. For instance, the templates for querying emotion are like ``What is the emotion in this face?'', ``What are the action units present in this face?''. More details about these templates are in the Appendix. 
FABA-Instruct statistics are shown in \cref{tab:stat}. For the face images, we randomly sample 19,474 and 403 face images from the training and testing set of AffectNet \cite{mollahosseini2017affectnet} as the training and testing face images, respectively.  AffectNet is a large-scale in-the-wild facial expression database, which crawls from the Internet by querying three search engines. We align and crop these face images using the Dlib library\footnote{\url{http://dlib.net/}}. 

However, some of the annotations for these face images cannot be obtained due to the low resolution and occlusion issues, especially for the AU annotations. As a result, after filtering out the useless annotations, we obtain 19,474 and 15,835 instructions in terms of emotions and AUs, respectively. Moreover, since some of the annotations are inaccurate or inexact, we carefully check and revise each annotation for the test sets of two tasks. 
\begin{figure}[t]
    \centering
    \includegraphics[width=0.95\textwidth]{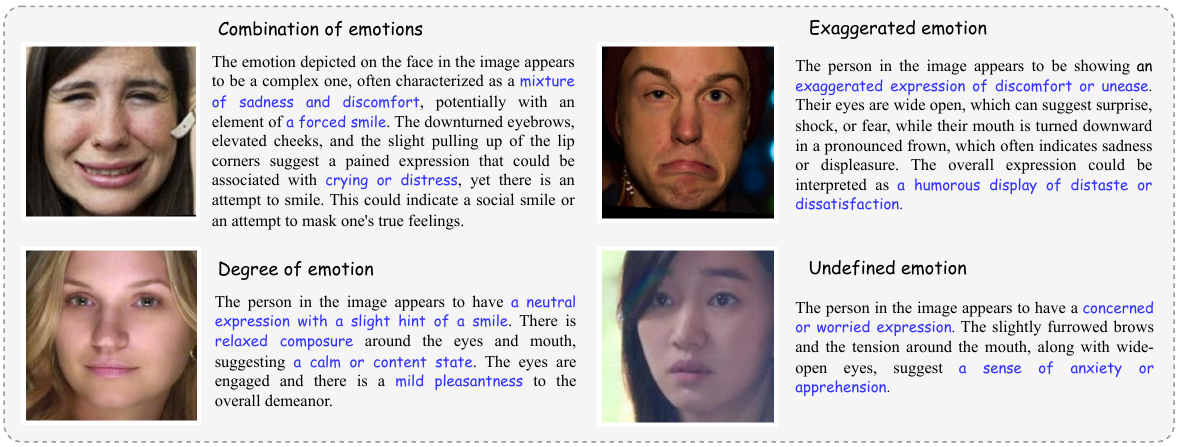}
    \caption{Emotion description analysis. Emotion descriptions can express comprehensive emotion types like compositional emotions, exaggerated emotions, the degree of emotions, and undefined emotions, \etc. In contrast, emotion categories struggle to capture such complex and nuanced emotional states.}
    \label{fig:emo_analysis}
\end{figure}

\noindent\textbf{Emotion description analysis.} Existing emotion datasets like RAF-DB \cite{li2017reliable}, AffectNet \cite{mollahosseini2017affectnet}, CK+ \cite{lucey2010extended}, FER2013 \cite{goodfellow2013challenges}, MMI \cite{pantic2005web}, SFEW \cite{dhall2011static}, mainly adopt the seven emotion categories, \ie, \texttt{happiness}, \texttt{sadness}, \texttt{anger}, \texttt{fear}, \texttt{disgust}, \texttt{surprise}, \texttt{neutral}. However, we argue that classifying emotions into one of the several discrete emotion categories is limited in practice
since emotions are subjective \cite{barrett2007experience}, complex \cite{ekman1978facial}, continuous \cite{liu2014learning}, and contextual \cite{haidt1999culture}. For instance, as shown in \cref{fig:emo_analysis}, it is inaccurate to classify the compositional expression, \texttt{sadness with a forced smile}, into one of the emotion categories. For some exaggerated emotions, \eg, \texttt{making a mouth}, the actual emotion is unknown from the face image 
and it is not reasonable to classify the emotion according to its superficial facial movements. Also, discrete emotions cannot express the degree of emotion, and it is inexact to classify the girl's face as either \texttt{happy} or \texttt{neutral}. Furthermore, since emotions are complex, basic emotions cannot cover undefined emotions like \texttt{worry}, \texttt{skeptical}, \etc. By contrast, the emotion descriptions can address these issues due to their expressive nature.

\begin{figure}[t]
    \centering
    \includegraphics[width=0.95\textwidth]{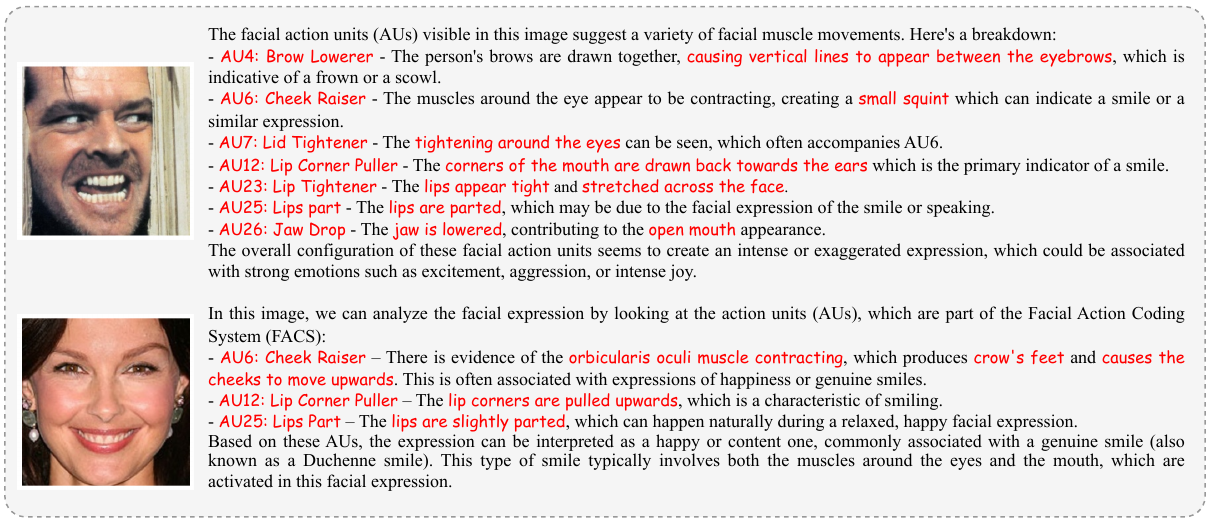}
    \caption{AU description analysis. AU descriptions give not only the AU labels, but also provide explanations on the cause (which muscle movement) and effect (which emotion it will lead to)  w.r.t. each AU, and the relationship between current AU and other AUs or emotions. }
    \label{fig:au_analysis}
\end{figure}

\noindent\textbf{Action Unit description analysis.} The annotations from traditional AUR datasets, \eg, BP4D \cite{zhang2014bp4d}, DISFA \cite{mavadati2013disfa}, EmotioNet \cite{fabian2016emotionet}, GFT \cite{girard2017sayette}, mainly adopt a string of binary vector to denote whether each AU is activated or not. However, the representation capability of this way is limited. It cannot indicate the degree of AU's activation and cannot provide any explanations and analysis for the prediction. For instance, in \cref{fig:au_analysis}, merely stating that AU6 (cheek raiser) is activated cannot depict the degree of cheek muscle raising. In contrast, the AU descriptions can provide more reasoning cues like ``\texttt{small squint}'' or ``\texttt{crow's feet}'', which can capture the activated degree of AU6. Furthermore, descriptions can also show the inference ability by providing the relationship between current AU and other AUs or emotions, \eg, ``\texttt{indicative of a frown or a scowl}'', ``\texttt{associated with strong emotions}''. Such descriptions can not only provide more nuances but also improve the interpretability \cite{li2023negative,chen2024less} of the model.

\subsection{Instruction-following FABA benchmark}\label{faba_bench}

\noindent\textbf{Evaluation.} Given the fact that our FABA-Instruct uses free-form textual descriptions to represent 
emotions and AUs, its model evaluation stands distinct from the traditional FABA tasks and Natural Language Generation (NLG) ~\cite{gatt2018survey} tasks. Specifically, for one thing, the traditional NLG metrics such as BLEU \cite{papineni2002bleu} or ROUGE \cite{lin2004rouge} scores mainly concentrate on the coherence and fluency of the generative text, ignoring the FABA relevant consideration in the evaluation.
For another, existing FABA metrics, \eg, accuracy or F1 score, focus only on the recognition performance of the model, lacking the evaluation on textual aspects of the model such as the reasoning and explanation. To compensate for the drawbacks of these two metrics, we introduce a new metric REGE to evaluate the \textbf{RE}cognition and \textbf{GE}neration performance of models on our FABA-Instruct.

Our REGE score is defined to consider both text generation and image recognition aspects of FABA models. For text generation, 
we choose to use the ROUGE score, which is a generally used metric in NLG by comparing the overlap of n-grams, word sequences, and word pairs between the generated texts and the reference ones. For recognition, we adopt the recognition accuracy for the multi-class performance of facial emotion recognition, and the F1 score for the multi-label performance of AU recognition. Denote the recognition performance as $S_{re}$ and the generation metric as $S_{ge}$, our REGE metric is computed by taking their sum: 
$S_{rege}=S_{re}+S_{ge}$.

\begin{figure}[t]
    \centering
    \includegraphics[width=0.95\textwidth]{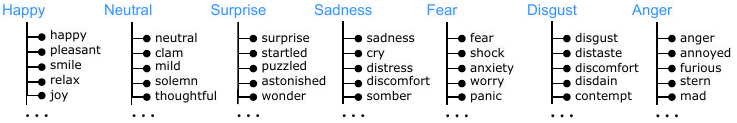}
    \caption{The synonyms of emotions for classifying the text.}
    \label{fig:emotion_synonyms}
\end{figure}

\begin{figure}[t]
    \centering
    \begin{subfigure}{0.48\textwidth}
        \centering
        \includegraphics[width=0.95\textwidth]{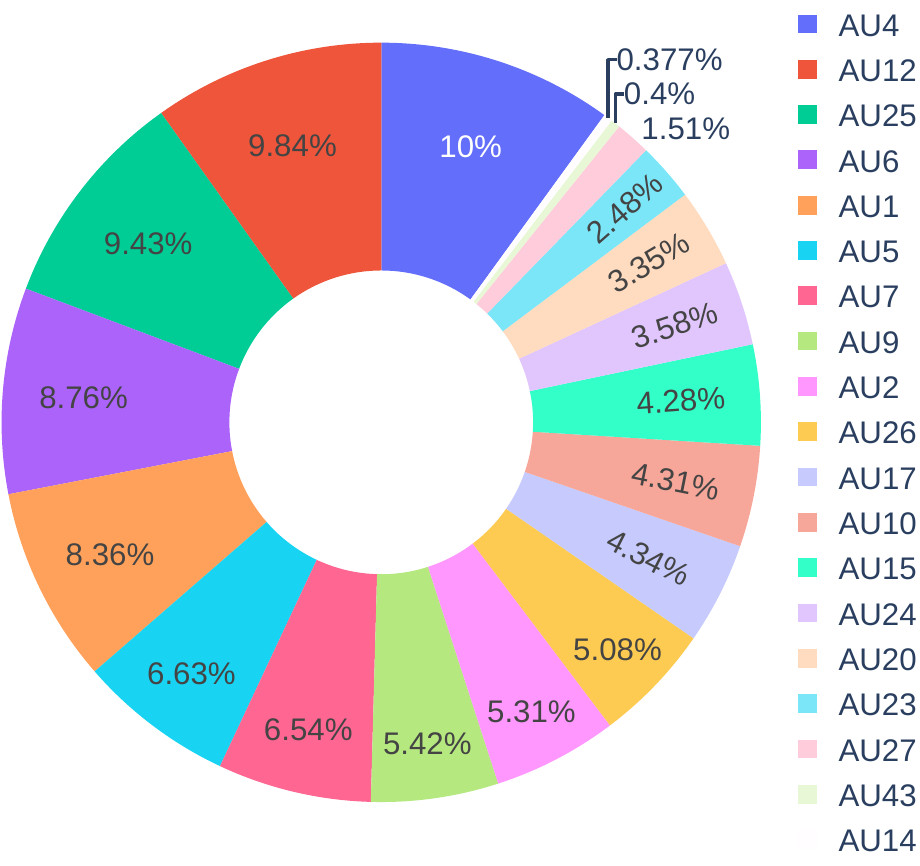}
        \caption{AU distribution.}\label{au_dist}
    \end{subfigure}
    \hfill
    \begin{subfigure}{0.48\textwidth}
        \centering
        \includegraphics[width=0.8\textwidth]{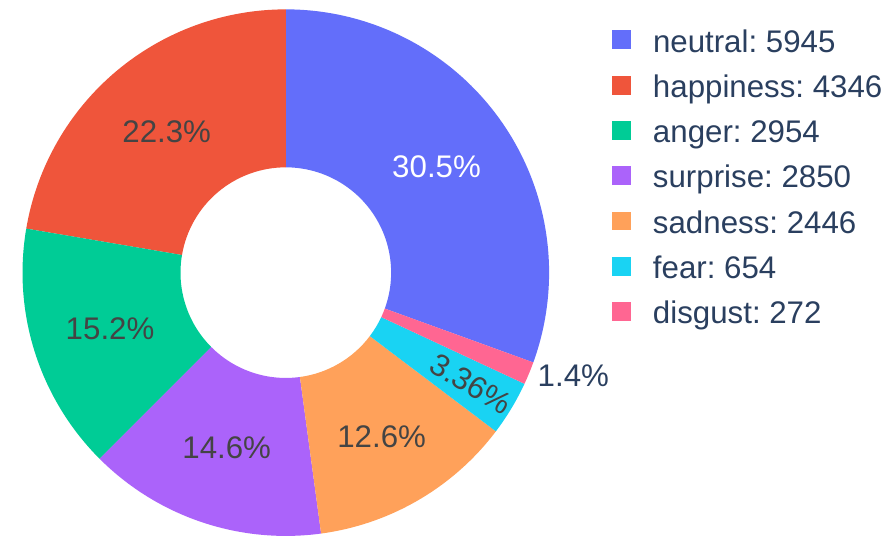}
        \caption{Emotion distribution.}\label{emo_dist}
        \centering
        \includegraphics[width=0.9\textwidth]{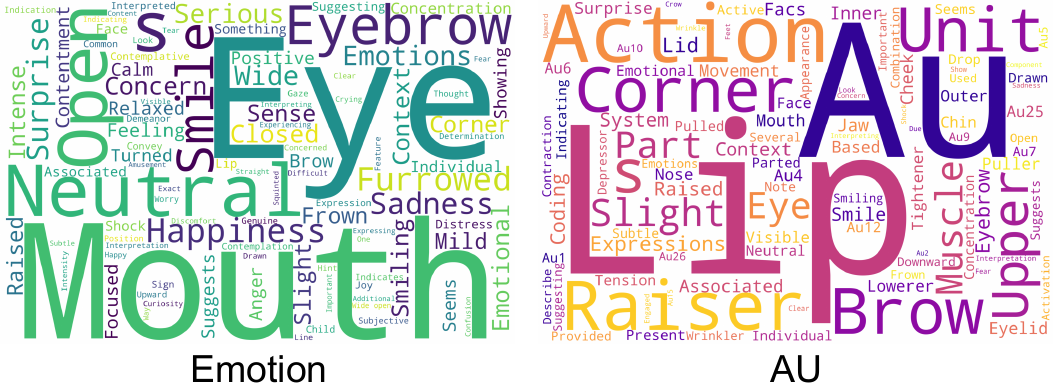}
        \caption{Word clouds.}\label{word_cloud}        
    \end{subfigure}
    \caption{The distribution of FABA-Instruct on AU (a) and emotion (b) tasks, and the word clouds (c). We extract the emotion labels using the synonyms of each emotion.}
    \label{fig:enter-label}
\end{figure}
\noindent\textbf{Calculation of $S_{re}$.} For the FER task, we propose to classify a face image 
into one of the aforementioned seven basic emotion categories. 
In order to classify the description, we manually select some synonyms from the training descriptions for each emotion category (see \cref{fig:emotion_synonyms}), and the emotion distribution of the training descriptions is shown in \cref{emo_dist}. In practice, since the descriptions incorporate negative sentences, we first deleted these sentences. After that, we count the frequency of emotion synonyms in the sentences and treat the emotion with the highest count as the emotion label for this description. After obtaining these \textit{emotion labels} of texts, we can calculate the accuracy score to evaluate the recognition performance on the FER task. 

Similarly, for the AUR task, we propose to classify a face image 
into one or multiple AU categories. We show the distribution of AUs in \cref{au_dist}. Following existing literature~\cite{zhang2014bp4d,fabian2016emotionet}, we choose 12 AUs for evaluation. 
Subsequently, we can calculate the F1 scores to represent the recognition performance for AUR.

\section{EmoLA: An Instruction-tuned MLLM for FABA}

\begin{wrapfigure}[11]{R}{0.65\linewidth}
    \centering
    \vspace{-20pt}
    \includegraphics[width=1\linewidth]{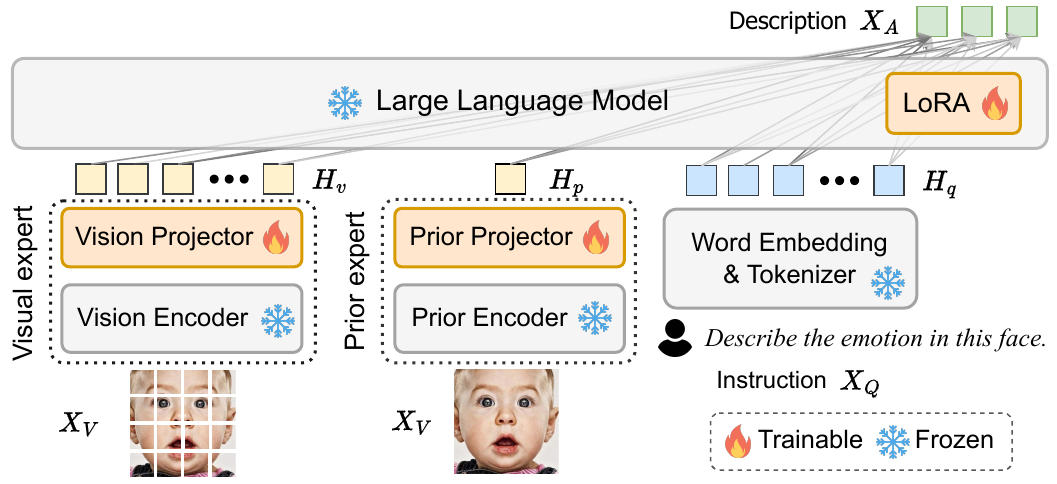}
    \captionsetup{font=small,aboveskip=3pt}
    \caption{EmoLA architecture.}
    \label{fig:emola}
\end{wrapfigure}

In this Section, we introduce a novel MLLM designed for FABA tasks. The overall framework of EmoLA is illustrated in \cref{fig:emola}. Its important components include two image experts (a visual and a facial prior expert), a language expert (a tokenizer with the word embedding) and a language decoder (LLM) with a LoRA module. Specifically, a face image $X_V$ is encoded by a visual expert known as a pre-trained CLIP-L/14~\cite{radford2021learning} with a two-layer Multi-Layer Perceptron (MLP), which generates the visual embedding tokens $H_v$. Similarly, for the input instruction $X_Q$, the language expert can provide the language tokens $H_q$. 

Notably, the visual tokens $H_v$ may fail to capture the facial structure information since CLIP is trained with general image-text pairs rather than FABA datasets, which may make the visual expert focus more on general semantic information and overlook task-specific features. Hence, we suggest employing an extra facial prior encoder $f_p(\cdot)$, trained on facial-related datasets, to better capture the facial prior knowledge and enhance the recognition ability for face images. Specifically, we adopt a pretrained facial landmark detector from Insightface\footnote{\url{https://github.com/deepinsight/insightface/tree/master}} to extract the landmark feature. Other face priors, such as recognition features  \cite{deng2019arcface}, parsing, \etc, can also be considered. We leave exploring these additional priors as future work for EmoLA. Therefore, the prior feature extracted by $f_p$ can be expressed as:
$Z_P = f_p(X_V).$
After obtaining $Z_P$, we utilize the MLP $g_{\theta}(\cdot)$ to project the facial prior feature $Z_P$ to the token embedding space:
\begin{equation}
    H_p = g_{\theta}(Z_P),
\end{equation}
where $H_p$ is the facial prior token. This token can provide prior knowledge of face structure, which may be ignored by the visual expert.

After obtaining the visual embedding tokens $H_v$, facial prior token $H_p$ and the language embedding tokens $H_q$, we concatenate them together and feed them into the LLM decoder. Here we use Vicuna \cite{vicuna2023} as the LLM decoder. As shown in \cref{fig:emola}, the visual encoder, prior encoder, word embedding and the LLM decoder are frozen. Except for the prior encoder $f_p(\cdot)$, the initial weights of the other modules come from a pretrained MLLM. In this paper, we utilize the LLaVA-1.5 \cite{liu2023improved} as the backbone model. In order to train efficiently without tuning the entire MLLM, we propose to tune an extra LoRA module $h_\phi(\cdot)$, a visual projector $h_{\gamma}(\cdot)$, and the prior projector $g_\theta(\cdot)$. Compared to finetuning the entire LLM, tuning LoRA reduces both memory and computation costs during training without inducing additional expenses for inference. Our experiments demonstrate the effectiveness of such a design in \cref{tab:FABA_bench}. As a result, the overall parameters to be optimized are $\boldsymbol{\Theta} = \{\theta, \gamma, \phi\}$. We optimize these parameters following the auto-regressive way, and the likelihood of the generated description $X_A$ conditioned on image $X_V$, prior facial feature $Z_{P}$ and instructions $X_{Q}$ is given by:
\begin{equation}
    p\left( X_A|X_V,Z_P,X_Q \right) =\prod_{i=1}^L{p_{\boldsymbol{\Theta}}\left( x_i|X_V,Z_P,X_{Q}, X_{A, <i} \right)},
\end{equation}
where $L$ is the length of the token sequence, $x_i$ is the current token that needs to be predicted. $X_{A, <i}$  are the previous answer tokens.

\begin{wraptable}[12]{r}{0.43\textwidth}\vspace{-45pt}
\setlength{\tabcolsep}{0.35mm}
{
  \begin{table}[H]
    \renewcommand\arraystretch{1}
    \centering
   \caption{Comparison on RAF-DB.}\vspace{-3pt}
   \label{tab:raf}
   \scalebox{0.87}{
    \begin{tabular}{cc}
      \toprule
      Methods & Accuracy (\%) \\
      \midrule
      RAN \cite{wang2020region} & 86.9 \\
      MA-Net \cite{zhao2021learning} & 87.22 \\
      EfficientFace \cite{zhao2021robust} & 88.36 \\
      RUL \cite{zhang2021relative} & 88.98 \\
      DAN \cite{wen2023distract} & 89.70 \\
      EAC \cite{zhang2022learn} & 90.35 \\
      APViT \cite{xue2022vision} & 91.98 \\
      POSTER \cite{mao2023poster} & \textbf{92.05} \\
      \rowcolor{gray!30} EmoLA (\textbf{Ours.}) & \textbf{92.05} \\
      \bottomrule
    \end{tabular}}
    \end{table}
}
\end{wraptable}
\vspace{-10pt}
\section{Experiments}
\vspace{-10pt}

\textbf{Implementation details.} We initialize all the frozen weights of EmoLA with LLaVA-1.5 7b \cite{liu2023improved}, and we only tune the prior projector and LoRA during the training stage. We train our EmoLA for one epoch optimized by AdamW with an initial learning rate of 1e-4 for all the datasets. The rank of LoRA is set to 128. We conduct all the experiments on 8 A6000 GPUs.

\noindent\textbf{Database and protocols.} We perform experiments on four traditional FER and AUR datasets, including one emotion dataset (RAF-DB \cite{li2017reliable}) and three AU datasets (BP4D \cite{zhang2014bp4d}, DISFA \cite{mavadati2013disfa} and GFT \cite{girard2017sayette}). 
We turn the annotations of these datasets into instruction-following ones by adding instructions. Instead of predicting the class index or binary vectors in discriminative models, our EmoLA directly outputs the corresponding emotions (\ie, ``Happy'') or AU labels (\ie, ``AU1, AU4''). We adopt accuracy and F1 score to evaluate the recognition performance for FER and AUR tasks, respectively. More details about these datasets can be found in the Appendix.  For BP4D and DISFA, following \cite{shao2021jaa,niu2019local,li2023recot}, we also perform a subject-exclusive 3-fold cross-validation. For GFT and RAF-FB, we train and test according to the original dataset division protocol. 

We also perform experiments on our FABA-Instruct dataset w.r.t. FER and AUR tasks. We compare with other MLLMs using our  SEGE metric (\cref{faba_bench}). We train and test all the models according to our dataset division protocol.

\subsection{Comparison on traditional FER and AUR datasets}
\setlength{\tabcolsep}{1.6mm}
{ 

  \begin{table}[t]
    \renewcommand\arraystretch{1}
    \centering
    \caption{F1 score (in $\%$) of 8 AUs on DISFA. All the results are from original papers.} \label{DISFA} \vspace{0pt}
    \scalebox{0.8}{
    \begin{tabular}{lccccccccc}
      \toprule
      Method/AU   &    1    & 2    & 4    & 6    & 9    & 12    & 25   & 26   & Avg.  \\
      \midrule
      DRML  \cite{zhao2016deep}   & 17.3 & 17.7 & 37.4 & 29.0 & 10.7 & 37.7 & 38.5 & 20.1 & 26.7 \\
      EAC-Net  \cite{li2018eac} & 41.5 & 26.4 & 66.4 & 50.7 &  {80.5} &  {89.3} & {88.9} & 15.6 & 48.5 \\
      DSIN \cite{corneanu2018deep}  & 42.4 & 39.0 & 68.1 & 28.6 & 46.8 & 70.8 & 90.4 & 42.2 & 53.6 \\
      SRERL \cite{li2019semantic} & 45.7 & 47.8 & 59.6 & 47.1 & 45.6 & 73.5 & 84.3 & 43.6 & 55.9 \\
      LP-Net \cite{niu2019local} & 29.9 & 24.7 & {72.7} & 46.8 & 49.6 & 72.9 & 93.8 & 65.0 & 56.9 \\
      CMS \cite{sankaran2019representation} & 40.2 & 44.3 & 53.2 &  {57.1} & 50.3 & 73.5 & 81.1 & 59.7 & 57.4 \\
      ARL \cite{shao2019facial} & 43.9 & 42.1 & 63.6 & 41.8 & 40.0 & 76.2 & {95.2} & 66.8 & 58.7 \\
      {SEV-Net}  \cite{yang2021exploiting} &  {55.3}&  {53.1} & 61.5 & 53.6 & 38.2 & {71.6} &  {95.7} & {41.5} & 58.8 \\
      {HMP-PS} \cite{song2021hybrid} & 38.0 & {45.9} & 65.2 & 50.9 & {50.8} & {76.0} & 93.3 & {67.6} & 61.0 \\
      {ATCM} \cite{jacob2021facial} & 46.1 & {48.6} &  {72.8} & 56.7 & 50.0 & {72.1} & 90.8 & {55.4} & {{61.5}} \\
      {ReCoT} \cite{li2023recot} & {51.3} & {36.2} & {66.8} & {50.1} & {52.4} & {{78.8}} &  {95.3} & {{69.7}} &  {62.6} \\
      J$\hat{\rm{A}}$A-Net \cite{shao2021jaa} & {62.4} & {60.7} & 67.1 & 41.1 & 45.1 & 73.5 & 90.9 & {{67.4}} & {63.5} \\
      {PIAP} \cite{tang2021piap} & {50.2} & {51.8} & {71.9} & {50.6} & {54.5} & {{79.7}} &  {94.1} & {{57.2}} &  {63.8} \\
      GraphAU (R-50) \cite{luo2022learning} & 54.6 & 47.1 & 72.9 & 54.0 & 55.7 & 76.7 & 91.1 & 53.0 & 63.1 \\
      \rowcolor{gray!25} {EmoLA (\textbf{Ours.})} & {50.5} &  {56.9} &  {83.5} & {55.2} & {43.1} & {{80.1}} & {91.6} & {{60.0}} &  {\textbf{65.1}} \\
  \bottomrule 
  \end{tabular}}
\end{table}}

\textbf{Comparison on traditional FER datasets.} We conduct a comparative experiment with the latest state-of-the-art (SOTA) methods on RAF-DB, as presented in \cref{tab:raf}. Our EmoLA achieves the best results compared with previous SOTA methods.  The results demonstrate the significant potential of MLLMs in addressing the FER problem. It is worth noting that most of these methods are specifically tailored for FER tasks, while our EmoLA is easy to adapt to other tasks (\eg, AUR) due to the high flexibility brought by instruction tuning.

\noindent \textbf{Comparison on traditional AUR datasets.} \cref{BP4D}, \cref{DISFA} and \cref{GFT} show the comparison results of EmoLA with other SOTA methods on AUR datasets. It can be observed that EmoLA outperforms all the other SOTA methods on DISFA and GFT datasets, achieving a 1.3\% improvement over PIAP \cite{tang2021piap} on DISFA, and a 3.5\% increase over EmoCo \cite{sun2021emotion} on GFT. Additionally, it closely competes with ReCoT on BP4D, with a marginal gap of only 0.6\%. The gap stems from the benefits of the consistency regularization and co-training in ReCoT to the BP4D dataset. We can observe that EmoLA excels in multi-label classification tasks, potentially due to its strategy of exclusively predicting the positive labels, thereby mitigating the imbalanced issue caused by negative labels. 
\setlength{\tabcolsep}{0.8mm}
{ 
    \begin{table*}[t]
      \renewcommand\arraystretch{1.1}
      \centering
      \caption{F1 score (in $\%$) of 12 AUs on BP4D. The results with * are taken from \cite{li2023recot}. All the other results are taken directly from their original papers. } \label{BP4D} \vspace{0pt}
      \scalebox{0.8}{       
      \begin{tabular}{lccccccccccccc}
        \toprule
       {Method/AU} & 1    & 2   & 4   & 6    & 7    & 10    & 12    & 14   & 15   & 17   & 23   & 24   & Avg.  \\
        \midrule
        DRML  \cite{zhao2016deep} & 36.4      & 41.8      & 43.0 & 55.0 & 67.0 & 66.3 & 65.8 & 54.1 & 33.2 & 48.0 & 31.7 & 30.0 & 48.3 \\                                       
        EAC-Net \cite{li2018eac} & 39.0      & 35.2      & 48.6 & 76.1 & 72.9 & 81.9 & 86.2 & 58.8 & 37.5 & 59.1 & 35.9 & 35.8 & 55.9\\
        DSIN \cite{corneanu2018deep} & 51.7      & 40.4      & 56.0 & 76.1 & 73.5 & 79.9 & 85.4 & 62.7 & 37.3 & 62.9 & 38.8 & 41.6 & 58.9 \\
        CMS \cite{sankaran2019representation} & 49.1      & 44.1      & 50.3 & {79.2} & 74.7 & 80.9 &  {88.3} & {63.9} & 44.4 & 60.3 & 41.4 & 51.2 & 60.6 \\
        LP-Net \cite{li2019semantic}  & 43.4      & 38.0      & 54.2 & 77.1 & 76.7 & 83.8 & 87.2 & 63.3 & 45.3 & 60.5 & 48.1 & 54.2 & 61.0 \\
        ARL \cite{shao2019facial} & 45.8      & 39.8      & 55.1 & 75.7 & 77.2 & 82.3 & 86.6 & 58.8 & 47.6 & 62.1 & 47.4 & 55.4 & 61.1 \\
        J$\hat{\rm{A}}$A-Net* \cite{shao2021jaa} & {47.2}      & {41.6} & {49.1} & 77.2 & 77.5 & 82.9 & 85.8 & 63.4 & 50.8 & 62.5 & 47.2 & 52.7 & 61.5 \\
	SRERL \cite{li2019semantic} & 46.9      & 45.3      & 55.6 & 77.1 & 78.4 & 83.5 & 87.6 & 60.6 & {52.2} &  {63.9} & 47.1 & 53.3 & 62.9 \\
        HMP-PS \cite{song2021hybrid} & 53.1   & 46.1      & 56.0 & 76.5 & 76.9 & 82.1 & 86.4 & 64.8 & 51.5 & 63.0 & {{49.9}} & 54.5 & 63.4 \\  
        SEV-Net \cite{yang2021exploiting} &  {58.2}      &  {50.4}       & 58.3 &  {81.9} & 73.9 &  {87.8} & 87.5 & 61.6 &  {52.6} & 62.2 & {44.6} & 47.6 & 63.9 \\                         
        PIAP \cite{tang2021piap} & 54.2 & 47.1 & 54.0 & 79.0 & 78.2 & 86.3 & 89.5 & 66.1 & 49.7 & 63.2 & 49.9 & 52.0 & 64.1 \\
        {ATCM} \cite{jacob2021facial} & 51.7      & {49.3} &  {61.0} & {77.8} & {{79.5}} & {82.9} & 86.3 & {{67.6}} & 51.9 & {63.0}  & {43.7} & {{56.3}} &  {64.2} \\
         GraphAU (R-50) \cite{luo2022learning} & 53.7 & 46.9 & 59.0 & 78.5 & 80.0 & 84.4 & 87.8 & 67.3 & 52.5 & 63.2 & 50.6 & 52.4 & {64.7} \\
         {ReCoT*}  \cite{li2023recot} & 51.5 & {47.8}  &  58.9 & {79.2} &  {80.2} & {{84.9}} &  {88.4} & {61.6} &  {53.3} &  {64.6}  &  {51.8} &  {55.4} &  \textbf{64.8} \\
         \rowcolor{gray!25} {EmoLA (\textbf{Ours.})} & 57.4 & {{52.4}} &  {61.0} & 78.1 & 77.8 & 81.9 & 89.5 & 60.5 & 49.3 & 64.9  & 46.0 & 52.4 & {64.2} \\
    \bottomrule 
    \end{tabular}
      }
\end{table*}}

\setlength{\tabcolsep}{1mm}
{ 

  \begin{table}[t]
    \renewcommand\arraystretch{1}
    \centering
    \caption{F1 score (in $\%$) results of 10 AUs on GFT. The results with * are taken from \cite{sun2021emotion}. The others are taken directly from original papers. ``ft'' stands for finetune.} \label{GFT} \vspace{0pt}
    \scalebox{0.8}{
    \begin{tabular}{lccccccccccc}
      \toprule
      Method/AU   &    1    & 2    & 4    & 6    & 10    & 12    & 14    & 15   & 23   & 24   & Avg.  \\
      \midrule
      {EACNet} \cite{li2018eac} & {15.5} & {56.6} & {0.1} & {81.0} & {76.1} & {{84.0}} &  {0.1} & {38.5} &  {57.8} &  {51.2} &  {46.1} \\
      {TCAE} \cite{li2019self} & {43.9} & {49.5} & {6.3} & {71.0} & {76.2} & {{79.5}} &  {10.7} & {28.5} &  {34.5} &  {41.7} &  {44.2} \\
      {ARL} \cite{shao2019facial} & {51.9} & {45.9} & {13.7} & {79.2} & {75.5} & {{82.8}} &  {0.1} & {44.9} &  {59.2} &  {47.5} &  {50.1} \\
      {MoCo (ft)*} \cite{he2020momentum} & 45.3 & 48.2 & 20.3 & 80.7 & 78.8 & 78.1 & 22.6 & 46.0 & 53.9 & 50.3 & 52.4 \\
      {Temporal Ranking (ft)*} \cite{lu2020self} & 58.8 & 56.8 & 33.2 & 72.5 & 76.2 & 80.8 & 19.9 & 46.8 & 55.2 & 47.3 & 54.7\\
      {J$\hat{\rm{A}}$ANet} \cite{shao2021jaa} & {46.5} & {49.3} & {19.2} & {79.0} & {75.0} & {{84.8}} &  {44.1} & {33.5} &  {54.9} &  {50.7} &  {53.7} \\
      {EmoCo} \cite{sun2021emotion} & {65.9} & {55.9} & {40.7} & {83.1} & {75.1} & {{81.4}} &  {21.3} & {48.5} &  {58.0} &  {56.5} &  {58.6} \\
      \rowcolor{gray!25} {EmoLA (\textbf{Ours.})} & {69.8} & {59.1} & {52.8} & {85.3} & {73.0} & {{85.3}} &  {32.3} & {47.6} &  {63.1} &  {52.2} &  {\textbf{62.1}} \\
  \bottomrule 
  \end{tabular}}
\end{table}}

\setlength{\tabcolsep}{0.4mm}
{
\begin{table}[t]
 \renewcommand\arraystretch{1.1}
  \centering
    \centering
    \setlength{\belowcaptionskip}{-2mm}
    \captionsetup{aboveskip=-10pt}
    \setlength{\extrarowheight}{0.5mm}
    \caption{Comparison on FABA-bench. All the baselines are reproduced by their open-sourced codes. $S_{re}$ in emotion and AU tasks stands for accuracy and average F1 scores, respectively. $S_{ge}$ and $S_{rege}$ are ROUGE and our REGE score. All scores are in $\%$.}
    \label{tab:FABA_bench}
    \scalebox{0.8}{
    \begin{tabular}{cccc|ccccccccccccccc}
      \toprule
       \multirow{2}{*}{Methods} & \multicolumn{3}{c|}{Emotion} & \multicolumn{15}{c}{AU} \\
        & $S_{re}$ & $S_{ge}$ & $S_{rege}$ & 1 & 2 & 4 & 5 & 6 & 10 & 12 & 17 & 24 & 25 & 26 & 43 & $S_{re}$ & $S_{ge}$ & $S_{rege}$ \\
      \midrule
      MiniGPT4-v2 \cite{chen2023minigpt} & 58.2 & 19.6 & 77.8 & 47.9 & 35.5 & 42.3 & 32.7 & 29.2 & 6.6 & 10.3 & 0.0 & 2.5 & 0.1 & 0.0 & 0.0 & 17.9 & 19.9 & 37.8\\
      mPLUG-Owl2 \cite{ye2023mplug} & 53.6 & 28.4 & 82.0 & 72.3 & 17.5 & 75.2 & 54.2 & 75.6 & 0.0 & 13.0 & 0.0 & 0.0 & 3.9 & 18.2 & 0.0 & 27.5 & 28.2 & 55.7\\
      Shikra \cite{chen2023shikra} & 62.5 & 32.1 & 94.6 & 70.6 & 33.9 & 76.6 & 63.3 & 57.8 & 43.4 & 58.0 & 53.0 & 54.1 & 68.5 & 42.4 & 0.0 & 51.8 & 34.8 & 86.6\\
      LLaVA-1.5 \cite{liu2023improved} & 62.3 & 31.6 & 93.9 & 74.2 & 32.7 & 76.5 & 67.9 & 63.6 & 41.0 & 61.0 & 53.4 & 54.1 & 67.5 & 43.5 & 50.0 & 57.1 & 34.3 & 91.4\\
      \rowcolor{gray!25} EmoLA (\textbf{Ours.}) & 64.5 & 31.7 & \textbf{96.2} & 72.8 & 37.3 & 79.9 & 67.3 & 69.9 & 41.7 & 63.6 & 56.8 & 55.6 & 73.4 & 56.8 & 0.0 & 56.3 & 35.2 & \textbf{91.5} \\
      \bottomrule
    \end{tabular}}
\end{table}
}
\subsection{Comparison on FABA-Bench}
We compare with current MLLMs, \ie, LLaVA-1.5 \cite{liu2023improved}, MiniGPT4-V2 \cite{chen2023minigpt}, Shikra \cite{chen2023shikra}, and mPLUG-Owl2 \cite{ye2023mplug}, on our FABA-Bench in \cref{tab:FABA_bench}. These baselines mainly employ Vicuna \cite{vicuna2023} as LLM decoder and incorporate instruction tuning. We reproduce all the baselines using their open-sourced codes.  For fair comparison, all the models are trained for 1 epoch and train on two tasks individually. Except for our EmoLA, all the other baselines are finetuned based on the pretrained MLLMs. More details about baselines can refer to the Appendix.

It is noteworthy that EmoLA achieves the best results in two tasks with fewer tuning parameters compared to other MLLMs. EmoLA finetunes merely 10\% of the parameters compared to LLaVA-1.5, yet it achieves better performance on FABA-Bench, which can be attributed to two aspects. For one thing, EmoLA only finetunes LoRA which is much more efficient than tuning the entire LLM decoder. For another, EmoLA incorporates a facial prior expert to extract the facial structure knowledge. This feature compensates for the information overlooked by the visual encoder in FABA tasks. 
Moreover, we can also see that due to the strong capability of LLM, the language generation 
ability of Shikra, LLaVA-1.5 and EmoLA is comparable, making it unreasonable to evaluate these models merely based on the NLG metrics. Our metric, REGE, accounts for both the recognition and generation abilities of MLLMs in FABA, offering a more comprehensive evaluation of their performance.
\subsection{Ablation study}\label{ablation}\vspace{-5pt}
We explore the effectiveness of the facial landmark token in \cref{tab:prior_token}, and the tuning strategies in \cref{tab:tuning}. More ablations can be found in Appendix, \eg, multi-task training, the position of the prior token, \etc \cref{fig:case_study} shows the capability of EmoLA on FABA tasks, with additional examples available in the Appendix.

\begin{wraptable}[9]{r}{0.43\textwidth}\vspace{-42pt}
\setlength{\tabcolsep}{0.35mm}
  \centering
  \setlength{\tabcolsep}{1.3mm}
    \begin{table}[H]\renewcommand\arraystretch{1.1}
        \centering
        \vspace{-2mm}
        \setlength{\extrarowheight}{-1mm}
        \caption{Influence of prior token. $H_v$ and $H_p$ indicate visual tokens and the facial prior token, respectively.}
        \label{tab:prior_token}
        \scalebox{0.9}{
        \begin{tabular}{cccccccc}
          \toprule
          \multicolumn{2}{c}{Tokens} & \multicolumn{2}{c}{Emotion} & \multicolumn{2}{c}{AU} \\
          $H_v$ & $H_p$ & $S_{re}$ & $S_{ge}$ & $S_{re}$ & $S_{ge}$ \\
          \midrule
           & \ding{51} & 41.2 & 29.9 & 40.5 & 33.7 \\
          \ding{51} &  & 62.5 & 32.1 & 55.3 & 34.8\\
          \ding{51} & \ding{51} & 64.5 & 31.7 & 56.3 & 35.2\\
          \bottomrule
        \end{tabular}}
        \end{table}
\end{wraptable}
\noindent\textbf{Effectiveness of prior token.} As shown in \cref{tab:prior_token}, if we input only the facial prior token $H_p$ to EmoLA without visual tokens, there will be a significant decline in recognition performance. However, thanks to the powerful capability of LLM decoder, the generalization ability does not suffer greatly. Furthermore, even with just a single token fed into the decoder, the model retains a certain level of recognition capability. This demonstrates that the landmark prior contains representative features for FABA tasks.

Compared to merely inputting the prior token $H_p$, incorporating visual tokens $H_v$ will enhance the recognition performance since these visual tokens carry more detailed face information. Incorporating both visual tokens and prior tokens leads to further improvements with eligible performance dropping in text generation. This suggests that the face prior token supplies complementary information of visual tokens. We suppose this additional information includes facial structure details introduced by landmark facial prior extractor.

\begin{figure}[t]
    \centering
    \setlength{\belowcaptionskip}{-20pt}
    \includegraphics[width=0.94\textwidth]{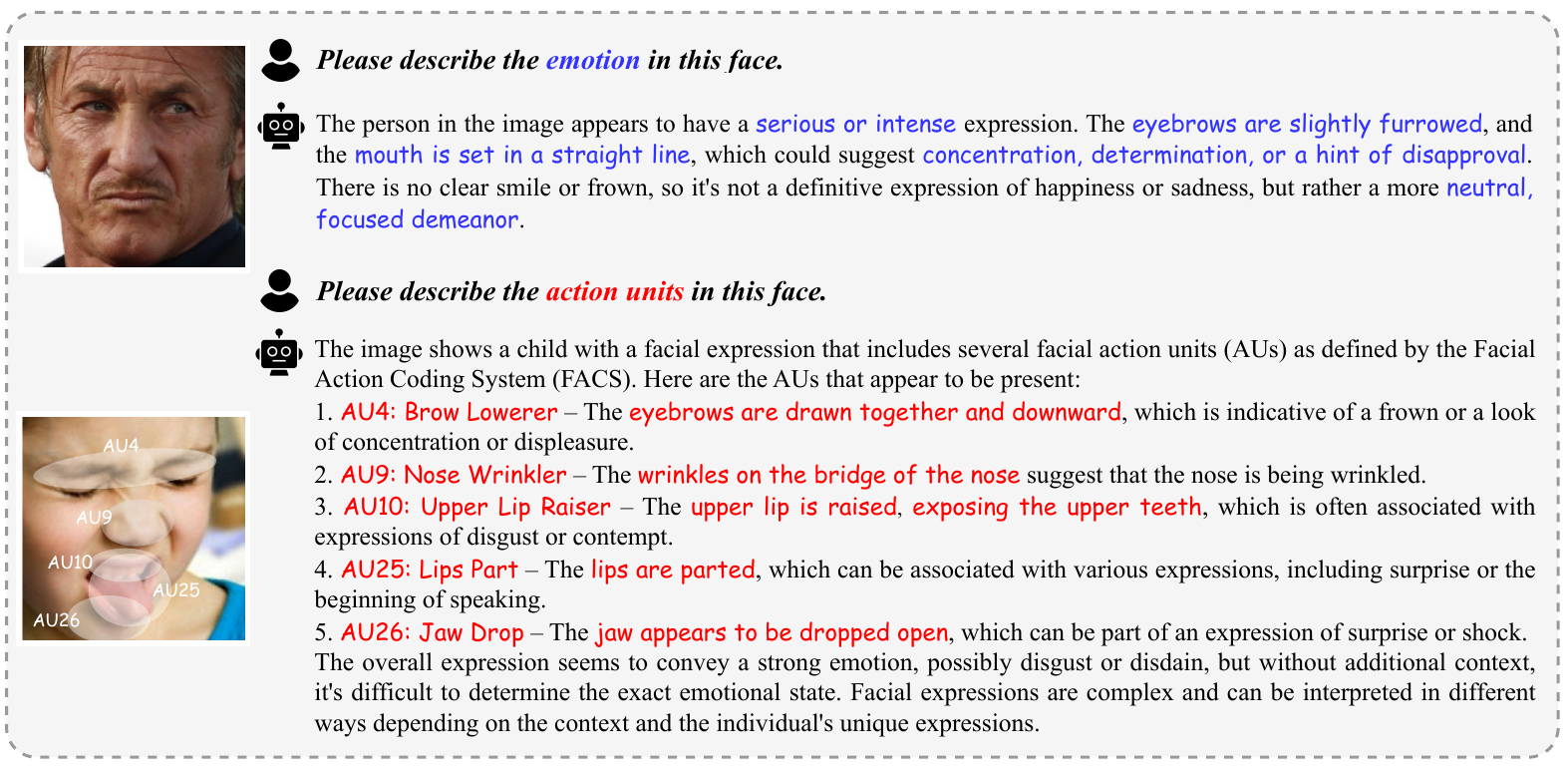}
    \caption{An example of EmoLA's capability on FABA tasks.}
    \label{fig:case_study}
\end{figure}

\noindent\textbf{Influence of tuning strategy.} We also investigate the impact of tuning strategies on two modules requiring optimization, $h_{\phi}$ and $g_{\theta}$, as shown in \cref{tab:tuning}. When only the prior projector $g_{\theta}$ is fine-tuned, with all other parameters kept frozen, we observe that while the overall recognition performance is relatively low, it still surpasses that achieved by using only a single prior token. This suggests the effectiveness of tuning the prior projector $g_{\theta}$, since the parameters of $g_{\theta}$ only take a small portion of the entire parameters.

\begin{wraptable}[10]{r}{0.42\textwidth}\vspace{-20pt}
\setlength{\tabcolsep}{1.2mm}
    \begin{table}[H]
        \centering
        \vspace{-2mm}
        \setlength{\extrarowheight}{-1mm}
        \caption{Influence of the tuning strategy. $h_{\phi}+h_{\gamma}$ and $g_{\theta}$ indicate the LoRA module with visual projector and the prior projector, respectively.}
        \label{tab:tuning}
            \scalebox{0.9}{
            \begin{tabular}{cccccc}
          \toprule
          \multicolumn{2}{c}{Modules} & \multicolumn{2}{c}{Emotion} & \multicolumn{2}{c}{AU}\\
          $h_{\phi}+h_{\gamma}$ & $g_{\theta}$ & $S_{re}$ & $S_{ge}$ & $S_{re}$ & $S_{ge}$ \\
          \midrule
          & \ding{51} & 44.9 & 29.6 & 47.7 & 34.0 \\
          \ding{51} &  & 63.0 & 32.1 & 55.6 & 34.9 \\
          \ding{51} & \ding{51} & 64.5 & 31.7 & 56.3 & 35.2\\
          \bottomrule
        \end{tabular}}
    \end{table}
\end{wraptable}
We also tried exclusively tuning LoRA with visual projector $h_{\phi}+h_{\gamma}$ while keeping other parameters frozen. This is different from merely inputting the visual tokens $H_v$ since it also includes a prior token $H_p$ generated by the randomly initialized prior projector $g_{\theta}$. The results indicate that tuning $h_{\phi}+h_{\gamma}$ significantly improves the performance by better aligning the output of the LLM. It can also be observed that the performance slightly exceeds that achieved using only the visual tokens, which also emphasizes the value of the prior token. Again, the efficacy is further augmented by simultaneously fine-tuning both $h_{\theta}+h_{\gamma}$ and the prior projector $g_{\theta}$.

\vspace{-5pt}
\section{Conclusion}
\vspace{-10pt}
In this paper, to address the challenges raised in FABA tasks when employing MLLMs, we proposed an instruction-following FABA dataset \textit{FABA-Instruct} by means of GPT4. Based on this dataset, we introduced a benchmark \textit{FABA-Bench} to comprehensively evaluate the FABA models on instruction-following data. Furthermore, we presented an instruction-tuned MLLM  \textit{EmoLA} for FABA, which is efficient and effective by tuning LoRA on a pre-trained MLLM and incorporating a facial prior expert. Extensive experiments across four traditional FABA datasets and our FABA-Bench demonstrate the effectiveness of EmoLA. 
In the future, we intend to broaden our method to additional facial-related tasks, \eg, face detection, face generation, \etc Moreover, incorporating other facial prior features holds the potential for performance improvement. Expanding our EmoLA from 2D face images to video streams also presents a promising avenue for future research.

\noindent\textbf{Acknowledgement}: Yifan Li, Wentao Bao, and Yu Kong are partially supported by NSF Awards 1949694 and 2040209. Any opinions, findings, and conclusions or recommendations expressed in this material are those of the authors and do not necessarily reflect the views of NSF. We also would like to express our deepest gratitude to Dr. Junwen Chen for her invaluable support and contribution to this research.
\bibliographystyle{splncs04}
\bibliography{main}

\clearpage
\appendix
\setcounter{section}{0}

\begin{center}
    \textbf{\Large Facial Affective Behavior Analysis with Instruction Tuning}
    \\ [0.8cm]
    {\Large Supplementary Material}
    \\ [1.2cm]
\end{center}

This is the supplementary material of Facial Affective Behavior Analysis with Instruction Tuning.

\section{Limitations and negative effects}
\subsection{Limitations}
This work also has its limitations. Firstly, we haven't tried other face-prior feature extractors except for landmark features. We leave this as an exploring research direction for future work. Secondly, our annotations from the training set also include noise induced by GPT-4V's hallucinations \cite{tan2024sparsity, tan2023interpreting} (see \cref{train_accuracy}), which may introduce some bias for models. We think our benchmark can be regarded as the learning from noisy descriptions task. Thirdly, the metric we proposed for evaluating the expression recognition ability may not reflect the nuanced expressions, which could be further improved.

\subsection{Negative effects}
There also exist some potential negative effects of our EmoLA. Firstly, \textit{privacy issues}. Facial affective behavior analysis (FABA) may infringe on users' privacy, especially when deploying EmoLA on public spaces or systems. Individuals' faces may be unconsciously captured and recorded, leading to privacy concerns. Secondly, \textit{misdirections and misjudgments}. Our EmoLA is not entirely accurate and may produce misjudged emotions. This can lead to misunderstanding or misdirection issues, especially in applications like security or judicial systems. Thirdly, \textit{technological abuse}. Our method may be abused to suppress dissent or monitor political activities, thereby leading to social hazard or freedom restrictions.

\section{Background of instruction tuning}
According to \cite{zhang2023instruction}, instruction tuning refers to ``the process of further training LLMs on a dataset consisting of (instruction, output) pairs in a supervised fashion, which bridges the gap between the next-word prediction objective of LLMs and the users' objective of having LLMs adhere to human instructions.'' The LLMs are typically trained on large general language corpora, which may differ from the users' objectives. As a result, to make the LLMs follow users' instructions, instruction tuning is proposed.  For instance, InstructGPT \cite{ouyang2022training}/ ChatGPT \cite{openai2023gpt}, FLAN-T5 \cite{flant5}, OPT-IML \cite{iyer2022opt} are tuned with instruction-following data to enable their counterparts GPT-3 \cite{brown2020language}, T5 \cite{raffel2020exploring}, OPT \cite{zhang2022opt} have better generalization and few-shot abilities. Inspired by the success of instruction tuning for LLMs,
LLaVA \cite{liu2024visual} attempts to extend this technique to the multimodal space, by introducing an MLP connector to map the visual tokens to language token space. Following that, other methods \cite{zhu2023minigpt, chen2023minigpt, chen2023shikra, ye2023mplug} also adopt this mechanism on multiple downstream tasks and achieve remarkable results. 

\section{Annotation details}
\subsection{Instructions in FABA-Instruct}
\begin{figure}[t]
    \begin{subfigure}{1\textwidth}        
        \centering
        \includegraphics[width=0.8\textwidth]{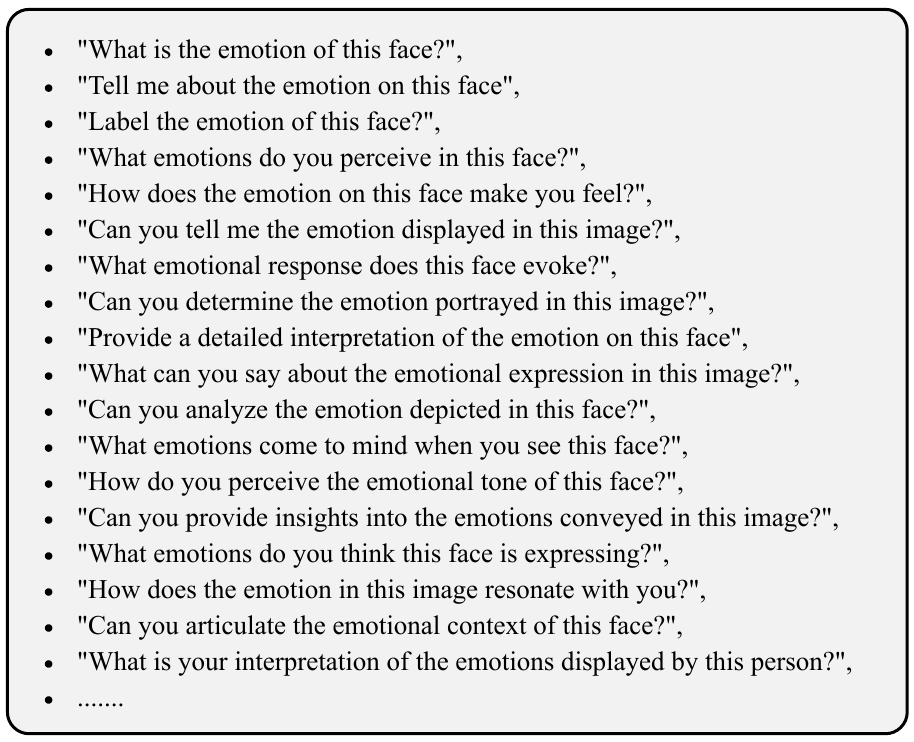}
        \caption{Examples of emotion instructions.}\vspace{10pt}
        \label{fig:emo_templates}
        \centering
        \includegraphics[width=0.8\textwidth]{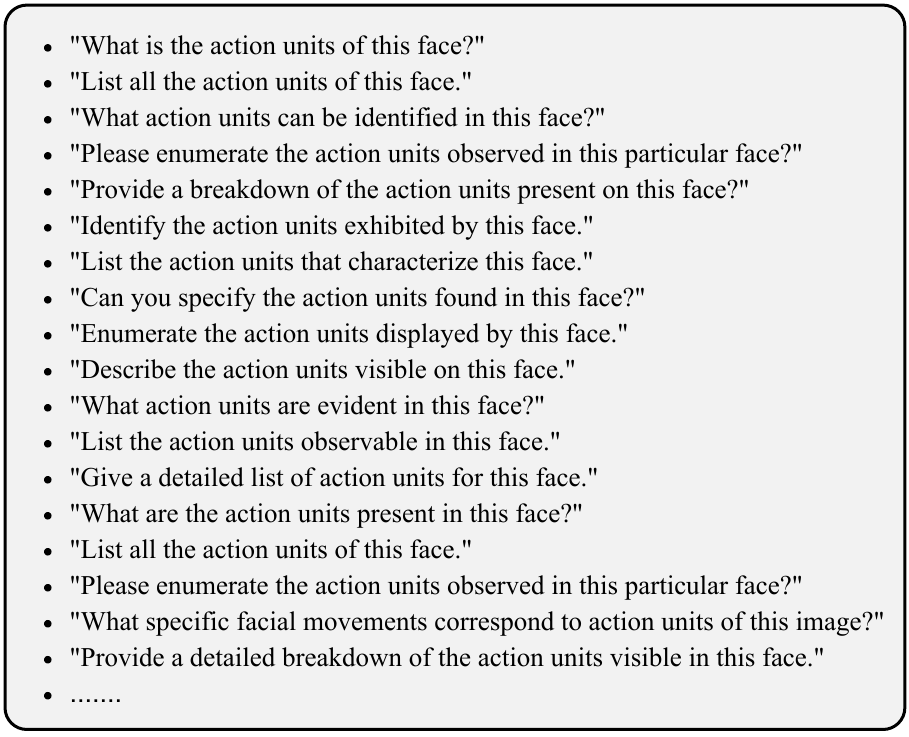}
        \caption{Examples of AU instructions.}
        \label{fig:au_templates}
    \end{subfigure}
    \caption{The instructions used in FABA-Instruct.}
\end{figure}

We tried different instructions in our FABA-Instruct datasets. Specifically, we adopt 100 carefully designed instructions for emotion and action unit (AU) recognition tasks, respectively. Some of these instructions for emotion and AU are delineated in \cref{fig:emo_templates} and \cref{fig:au_templates}, respectively. As shown in these examples, these instructions are all with natural language format.

\subsection{AU types in FABA-Instruct}
As mentioned in the paper, we select 12 AUs of FABA-Instruct for evaluation, \ie, AU1, AU2, AU4, AU5, AU6, AU10, AU12, AU17, AU24, AU25, AU26, AU43. Also, there exists in total of 19AUs in the training annotations, and the meaning of these AUs are given in \cref{tab:AU_meaning}.

\setlength{\tabcolsep}{4mm}
{
  \begin{table}[t]
    \renewcommand\arraystretch{1}
    \centering
   \caption{The meaning of AUs in FABA-Instruct.}
   \label{tab:AU_meaning}
    \begin{tabular}{cc}
      \toprule
      \textbf{AUs} & \textbf{Meaning} \\
      \midrule
      AU1 & inner brow raiser \\
      AU2 & outer brow raiser \\
      AU4 & brow lowerer \\
      AU5 & upper lid raiser \\
      AU6 & cheek raiser \\
      AU7 & lid tightener \\
      AU9 & nose wrinkler \\
      AU10 & upper lip raiser \\
      AU12 & lip corner puller \\
      AU14 & dimpler \\
      AU15 & lip corner depressor \\
      AU17 & chin raiser \\
      AU20 & lip stretcher \\
      AU23 & lip tightener \\
      AU24 & lip pressor \\
      AU25 & lips part \\
      AU26 & jaw drop \\
      AU27 & mouth stretch \\
      AU43 & eyes closed \\
      \bottomrule
    \end{tabular}
    \end{table}
}
\subsection{The accuracy estimation of training annotations}\label{train_accuracy}
There also exists some noise in the training set due to the hallucinations in GPT-4V. To estimate the label accuracy in training annotations, we randomly sample 200 samples from each task in FABA-Instruct, and manually re-annotate these samples. After that, we can roughly estimate the accuracy or F1 score of training annotations according to these manual annotations. Specifically, for the emotion task, we calculate accuracy by classifying the text into 7 classes, which has been introduced in the main content. For the AU task, we evaluate all the AUs using the F1 score.

\begin{figure}[t]
        \centering
    \begin{subfigure}{1\textwidth}        
        \centering
        \includegraphics[width=1\textwidth]{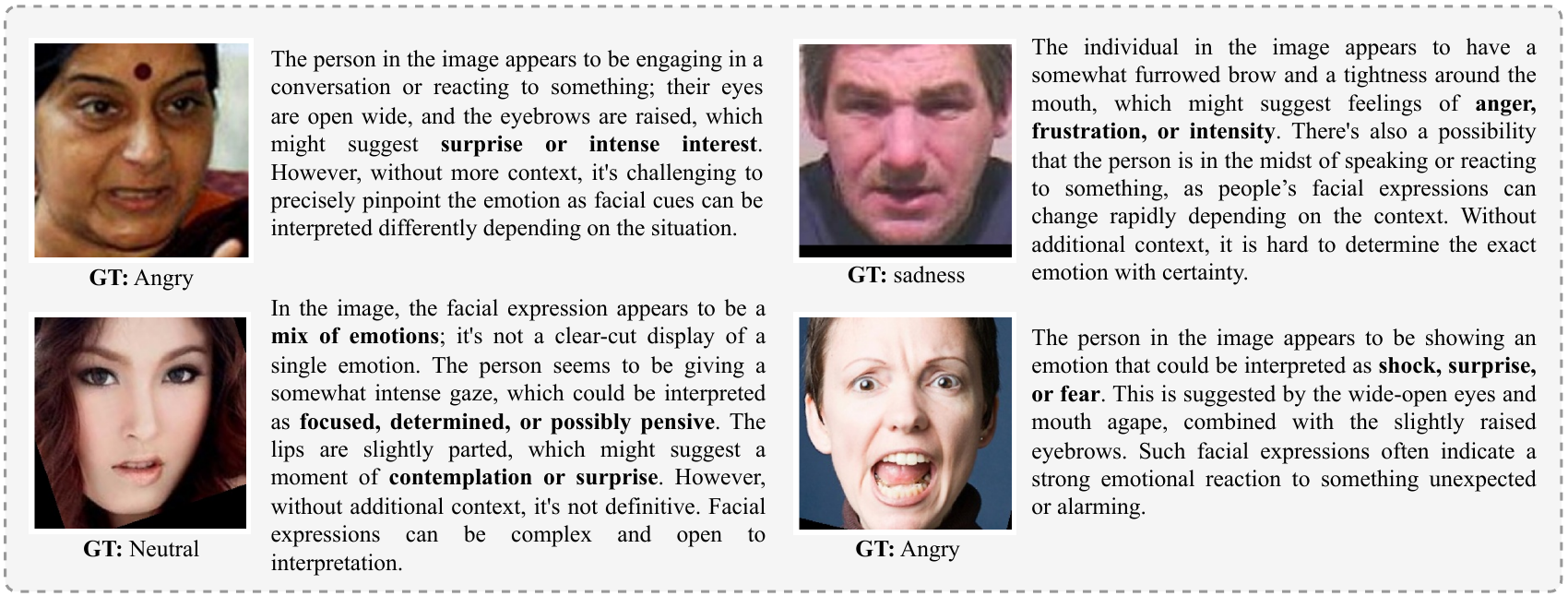}
        \caption{Examples of wrong emotion annotations.}
        \label{fig:emo_anno_fail}
        \vspace{10pt}
        \includegraphics[width=1\textwidth]{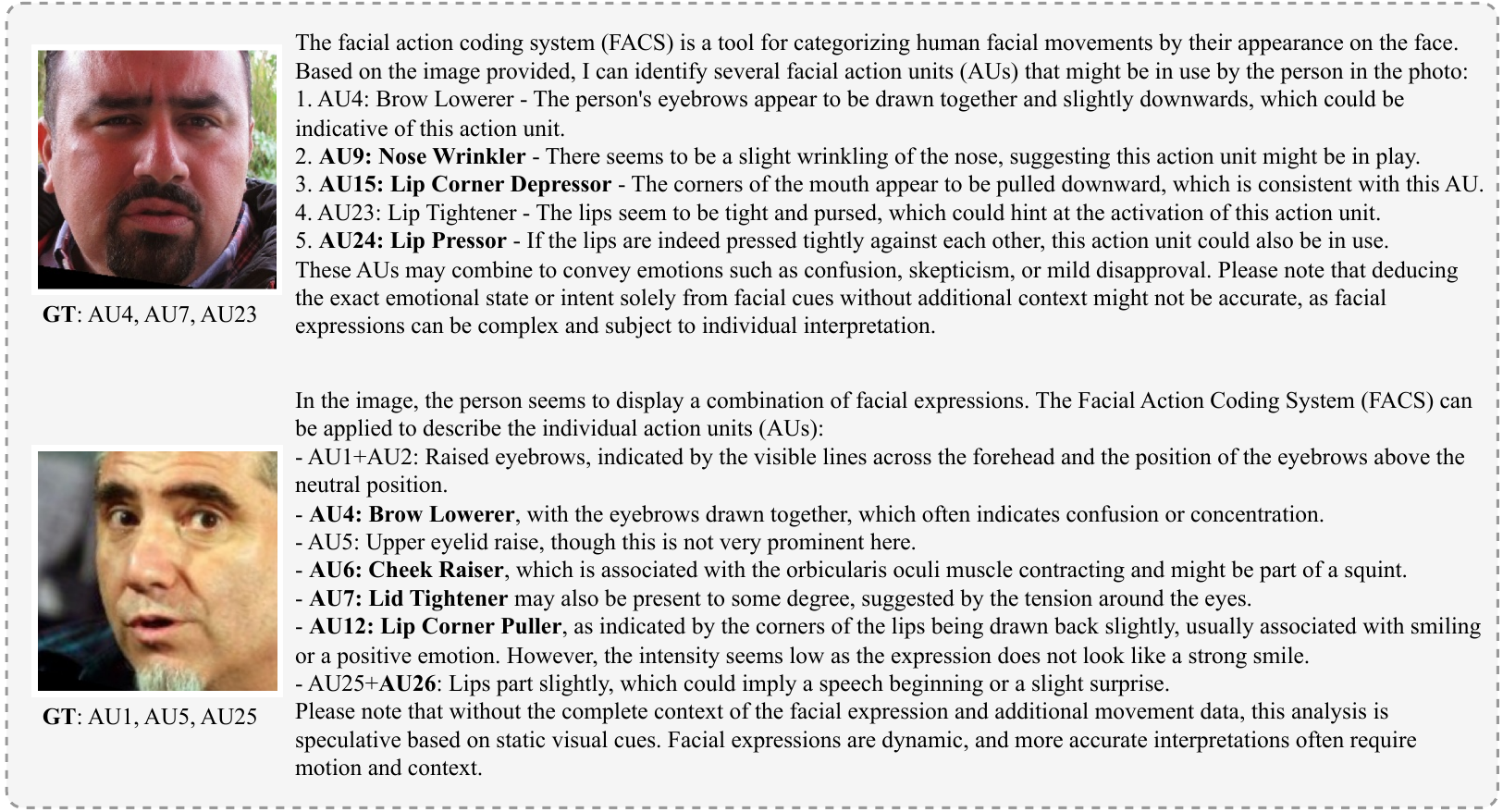}
        \caption{Examples of wrong AU annotations.}\vspace{10pt}
        \label{fig:au_anno_fail}
    \end{subfigure}
    \caption{Wrong annotation examples on FABA-Instruct.}\label{wrong_labels}
\end{figure}

For emotion annotations in FABA-Instruct, as shown in \cref{tab:emo_annos}, the accuracy of training annotations is about 91\%. For AU annotations, the average F1 is 76.1\% for all the AUs (see \cref{tab:AU_meaning}). From the estimation results, it can be observed that although there are some noisy labels in both two tasks, the recognition performance of GPT-4V on two tasks is still high. Therefore, it's reasonable to use these annotations for further research. Our FABA-Bench can not only be utilized for FABA tasks but also be regarded as the learning from noisy annotations task. Some examples are be found in \cref{wrong_labels}.

\setlength{\tabcolsep}{1.4mm}
{
\begin{table}[t]
  \centering
  \begin{minipage}[b]{1\textwidth}
    \begin{table}[H]
        \centering
        \vspace{-2mm}
        \setlength{\extrarowheight}{-1mm}
        \caption{Estimated accuracy of Emotion annotations in FABA-Instruct.}
        \label{tab:emo_annos}
        \begin{tabular}{ccc}
          \toprule
          Total & Correct & Acc. \\
          \midrule
          200 & 182 & 91.0 \\
          \bottomrule
        \end{tabular}
    \end{table}
    \vspace{-2mm}
    \begin{table}[H]
        \centering
        \setlength{\extrarowheight}{-1mm}
        \caption{Estimated F1 score of AU annotations in FABA-Instruct.}
        \label{tab:au_annos}
            \begin{tabular}{ccccccccccc}
          \toprule
          AU1 & AU2 & AU4 & AU5 & AU6 & AU7 & AU9 & AU10 & AU12 & AU15\\
          \midrule
          92.2 & 95.6 & 84.2 & 90.7 & 82.6 & 65.9 & 61.3 & 58.1 & 84.3 & 70.6 \\  
          \bottomrule
        \end{tabular}
        \vspace{8pt}
        \begin{tabular}{cccccccccc}
          \toprule
           AU17 & AU20 & AU23 & AU24 & AU25 & AU26 & AU27 & AU43 & \textbf{Avg.} \\
          \midrule
           66.7 & 55.2 & 79.1 & 79.5 & 84.9 & 83.9 & 78.8 & 57.1 & 76.1 \\  
          \bottomrule
        \end{tabular}
    \end{table}
  \end{minipage}
\end{table}
}
\section{Experiments}

\subsection{Details about the traditional FABA datasets}
We utilize three traditional AU datasets (BP4D, DISFA, GFT) and one traditional emotion datasets (RAF-DB) for better evaluating the performance of our EmoLA compared with the previous SOTA methods. Furthermore, we also sampled around 20,000 face images from AffectNet to construct our FABA-Instruct dataset.

\noindent \textbf{Emotion datasets}. \textit{AffectNet} \cite{mollahosseini2017affectnet} is a large-scale in-the-wild emotion database with more than 1M face images crawled from internet. We randomly sampled 20,000 images from this database for dataset construction. RAF-DB \cite{li2017reliable} is an facial expression dataset with 29,672 in-the-wild face images, which is annotated by 40 annotators with 7 single-label and two-tab compond emotion categories. In this paper, we mainly compare with other methods on the single-label subset.

\noindent \textbf{AU datasets}. \textit{BP4D} \cite{zhang2014bp4d} is a spontaneous facial AU dataset with 328 videos from 41 subjects (23 females and 18 males). There are in total of 140,000 frames with 12 AUs (1, 2, 4, 6, 7, 10, 12, 14, 15, 17, 23, 24). \textit{DISFA} \cite{mavadati2013disfa} consists of 26 subjects (12 females and 14 males) and 130,000 frames with the AU intensities (from 0 to 5) annotations. Following \cite{zhao2016deep}, an AU with the intensity equal or greater than 2 is considered to be activated. 8 (1, 2, 4, 6, 9, 12, 25, 26) of the 12 AUs are utilized for evaluation. \textit{GFT} \cite{girard2017sayette} contains 96 participants from 32 three-person groups. 10 AUs (1, 2, 4, 6, 10, 12, 14, 15, 23, 24) are selected considering the challenges brought by head motion and occlusion. There are in total of 108000 training images and 24600 evaluating images in GFT.

\subsection{Baseline details of FABA-Bench}
As presented in the Experiment section, we reproduced four baselines which have similar arhitectures as our EmoLA on FABA-Bench, \ie, MiniGPT4-V2 \cite{chen2023minigpt}, mPLUG-Owl2 \cite{ye2023mplug}, Shikra \cite{chen2023shikra}, and LLaVA-1.5 \cite{liu2023improved}. LLaVA utilize a image encoder to obtain the visual tokens, and map the visual tokens to the language space through a linear layer. After that, the image tokens and language tokens are passed to an LLM decoder to generate descriptions. LLaVA undergoes a two-stage training process: initially, it exclusively trains the projector, followed by a phase where only the LLM decoder is trained. LLaVA-1.5 enhances its performance by incorporating a two-layer MLP and adopting higher image resolutions. Shikra shares a similar architecture with LLaVA; however, it distinguishes itself by fine-tuning both the projector and the LLM decoder during its training phases. Similarly, MiniGPT4-V2, while architecturally similar to LLaVA, employs higher-resolution images to improve visual perception and aggregates every four neighboring visual tokens into a single token to optimize training efficiency. mPLUG-Owl2 introduced a visual abstractor module which aggregates the information from visual tokens by learnable queries. Furthermore, mPLUG-Owl2 leverarged a modality-adaptive module for facilitating multi-modal inputs to a shared semantic space for enabling modality collaboration.
\setlength{\tabcolsep}{0.8mm}
{
\begin{table}[t]
 \renewcommand\arraystretch{1.1}
  \centering
    \centering
    \setlength{\belowcaptionskip}{-2mm}
    \captionsetup{aboveskip=-10pt}
    \setlength{\extrarowheight}{0.5mm}
    \caption{The multitask performance of EmoLA on FABA-Bench.}
    \label{tab:FABA-multitask}
    \scalebox{1}{
    \begin{tabular}{cccc|ccc}
      \toprule
       \multirow{2}{*}{Methods} & \multicolumn{3}{c|}{Emotion} & \multicolumn{3}{c}{AU} \\
        & $S_{re}$ & $S_{ge}$ & $S_{rege}$ & $S_{re}$ & $S_{ge}$ & $S_{rege}$ \\
      \midrule
      EmoLA (single task) & 64.5 & 31.7 & {96.2} & 56.3 & 35.2 & {91.5} \\
      EmoLA (multi-task) & 64.3 & 32.0 & {96.3} & 54.7 & 33.9 & {88.6} \\
      \bottomrule
    \end{tabular}}
\end{table}
}

\subsection{Multi-task performance on FABA-Bench}
We also perform experiments to evaluate the multi-task performance of EmoLA on our FABA-Bench. Specifically, we train our EmoLA using dataset from two tasks instead of the individual dataset.
As shown in \cref{tab:FABA-multitask}, the performance on Emotion of EmoLA under multi-task setting is almost the same to the single task setting. While for AU task, the performance of multi-task EmoLA drops compared to the single task version. We assume it's because the emotion recognition task will somehow affect the AU recognition performance by making model focus more on the general emotion recognition. Moreover, the multi-task is harder than the single task, which may also decrease the performance on AU task.

\setlength{\tabcolsep}{0.8mm}
{
\begin{table}[t]
 \renewcommand\arraystretch{1.1}
  \centering
    \centering
    \setlength{\belowcaptionskip}{-2mm}
    \captionsetup{aboveskip=-10pt}
    \setlength{\extrarowheight}{0.5mm}
    \caption{The location of prior token.}
    \label{tab:prior_loc}
    \scalebox{1}{
    \begin{tabular}{cccc|ccc}
      \toprule
       \multirow{2}{*}{Methods} & \multicolumn{3}{c|}{Emotion} & \multicolumn{3}{c}{AU} \\
        & $S_{re}$ & $S_{ge}$ & $S_{rege}$ & $S_{re}$ & $S_{ge}$ & $S_{rege}$ \\
      \midrule
      EmoLA (prior token before visual tokens) & 63.5 & 32.1 & {95.6} & 55.4 & 34.3 & {89.7} \\
      EmoLA (prior token after visual tokens) & 64.5 & 31.7 & {96.2} & 56.3 & 35.2 & {91.5} \\
      \bottomrule
    \end{tabular}}
\end{table}
}
\subsection{Location of prior token}
We also investigate the location of prior token in EmoLA in \cref{tab:prior_loc}. From the results, we can observe that if we put the prior token before the visual tokens the recognition performance on two tasks will decrease. We assume this will affect the visual tokens' representation due to the causal mask in the decoder. As a result, we put the prior token after the visual tokens.

\section{More generation cases}

\subsection{Successful cases}
\begin{figure}[t]
        \centering
        \includegraphics[width=1\textwidth]{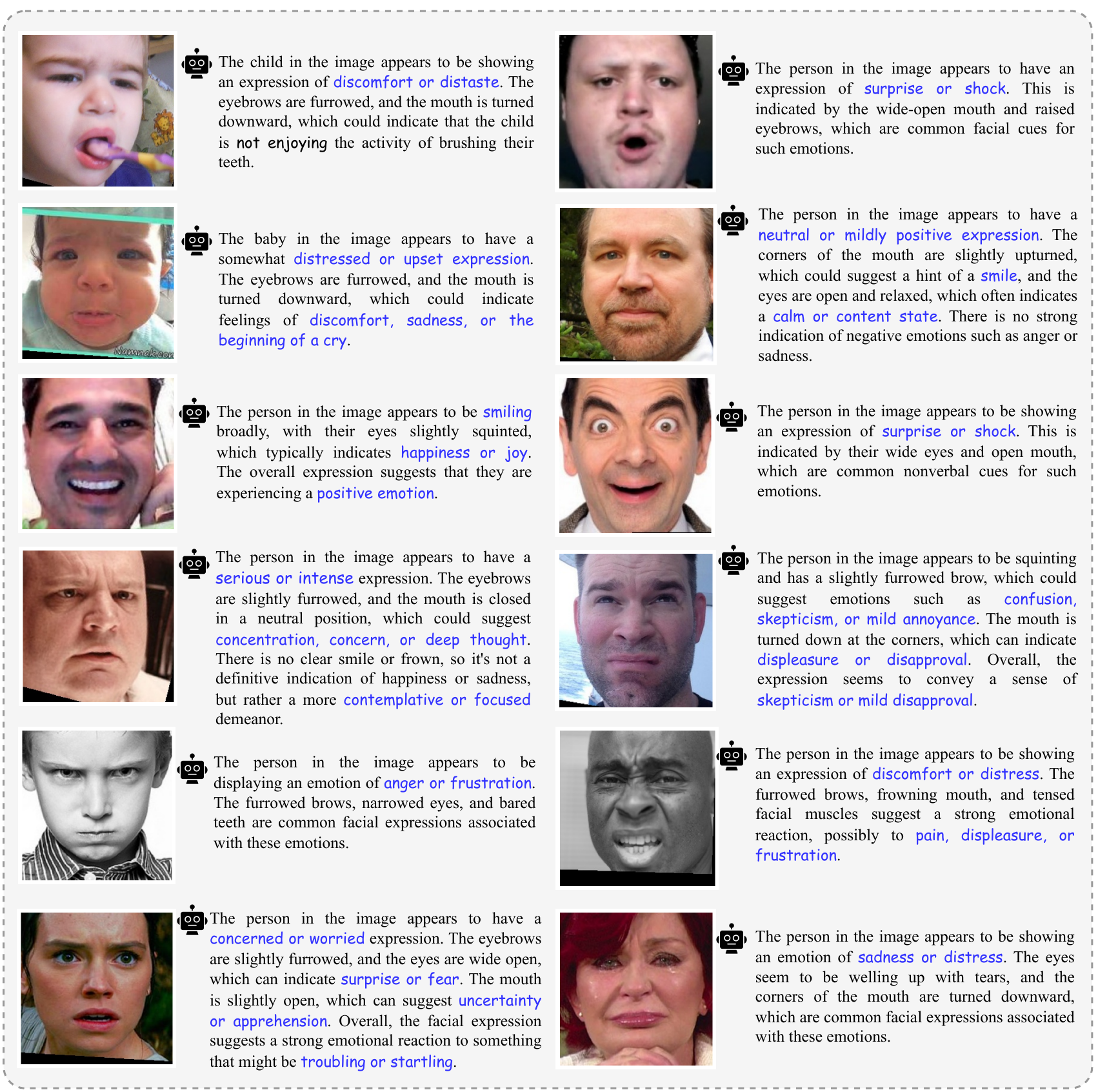}
        \caption{Examples of EmoLA's successful prediction on emotion task.}\vspace{10pt}
        \label{fig:emo_success}
\end{figure}

\begin{figure}[t]
        \centering
        \includegraphics[width=1\textwidth]{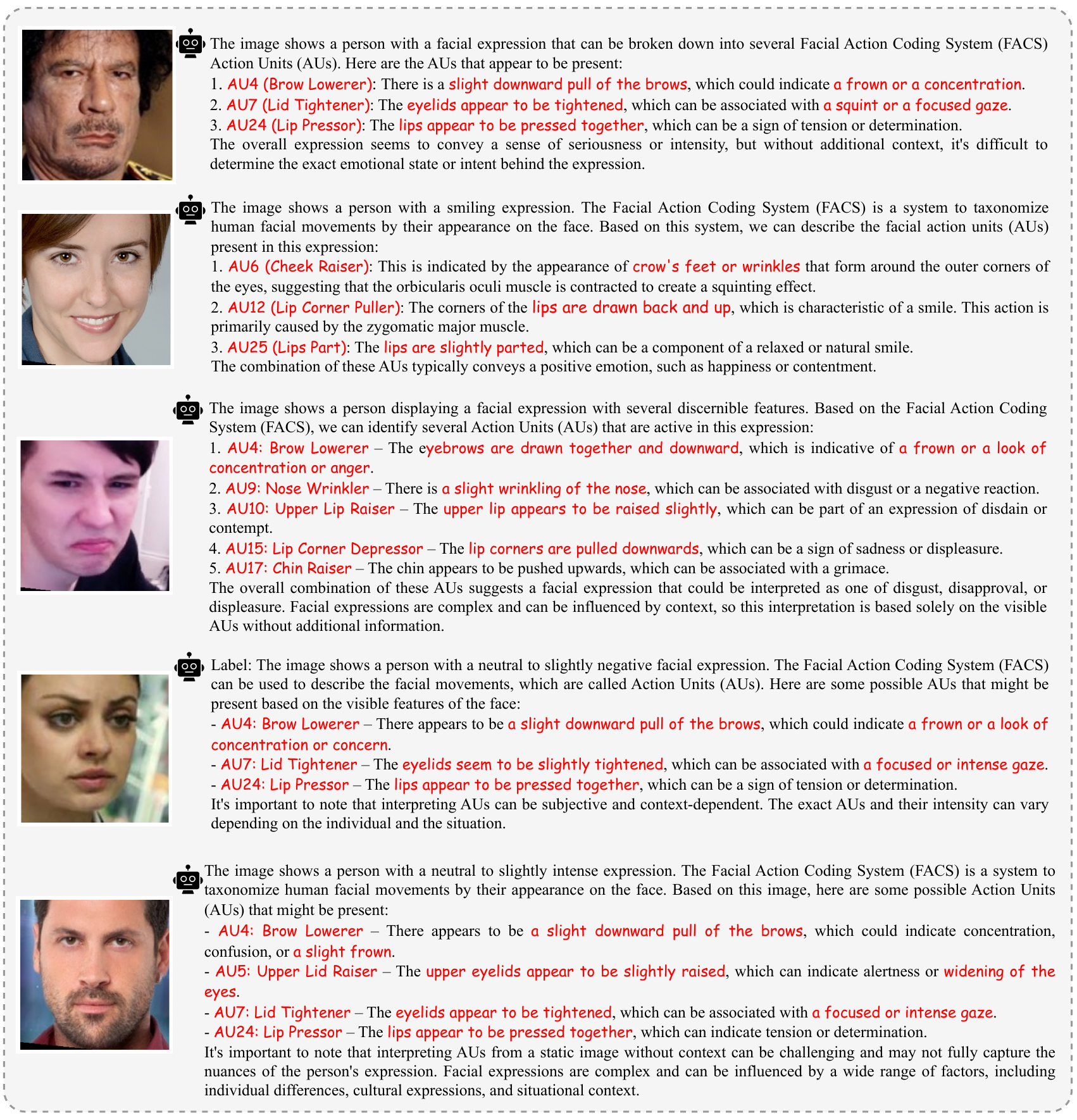}
        \caption{Examples of EmoLA's successful prediction on AU task.}
        \label{fig:au_success}
\end{figure}

We present some successful cases of EmoLA's prediction on our FABA-Instruct dataset in \cref{fig:emo_success} and \cref{fig:au_success}, respectively.


\subsection{Failed cases}
\begin{figure}[t]
        \centering
    \begin{subfigure}{1\textwidth}        
        \centering
        \includegraphics[width=1\textwidth]{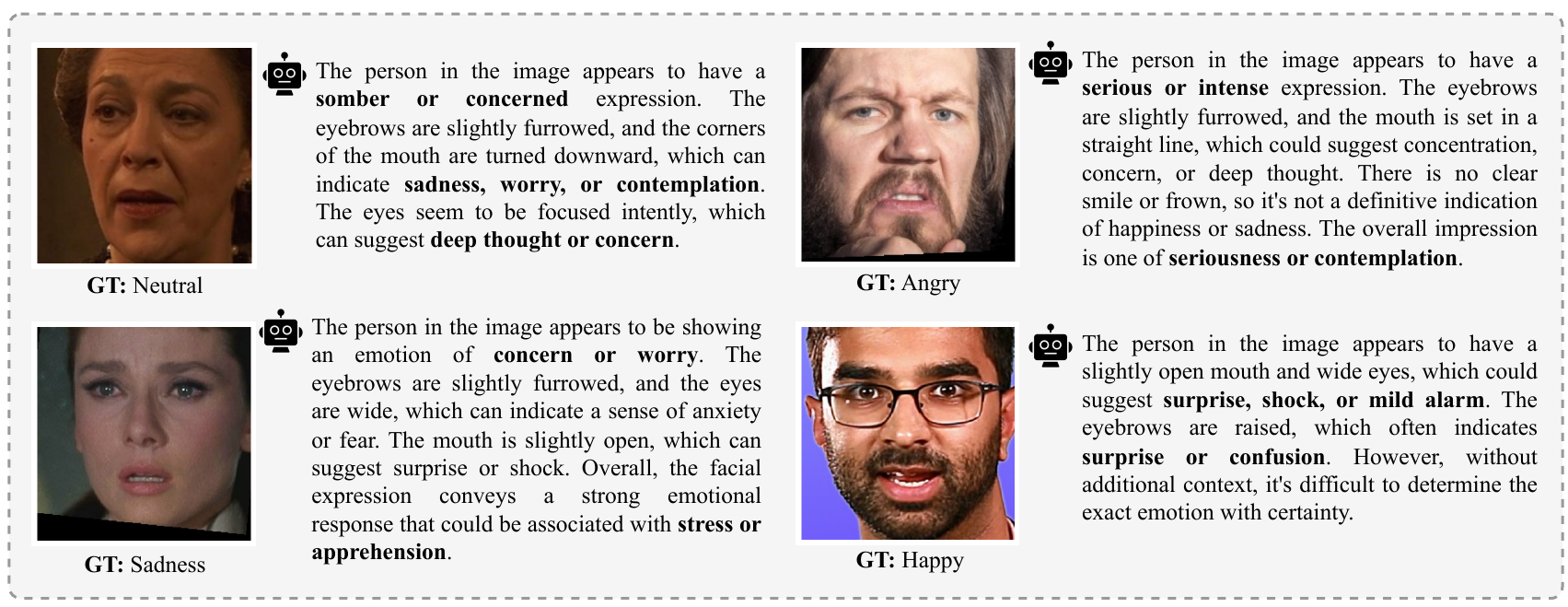}
        \caption{Examples of EmoLA's failed prediction on emotion task.}
        \label{fig:emo_fail}
        \vspace{10pt}
        \includegraphics[width=1\textwidth]{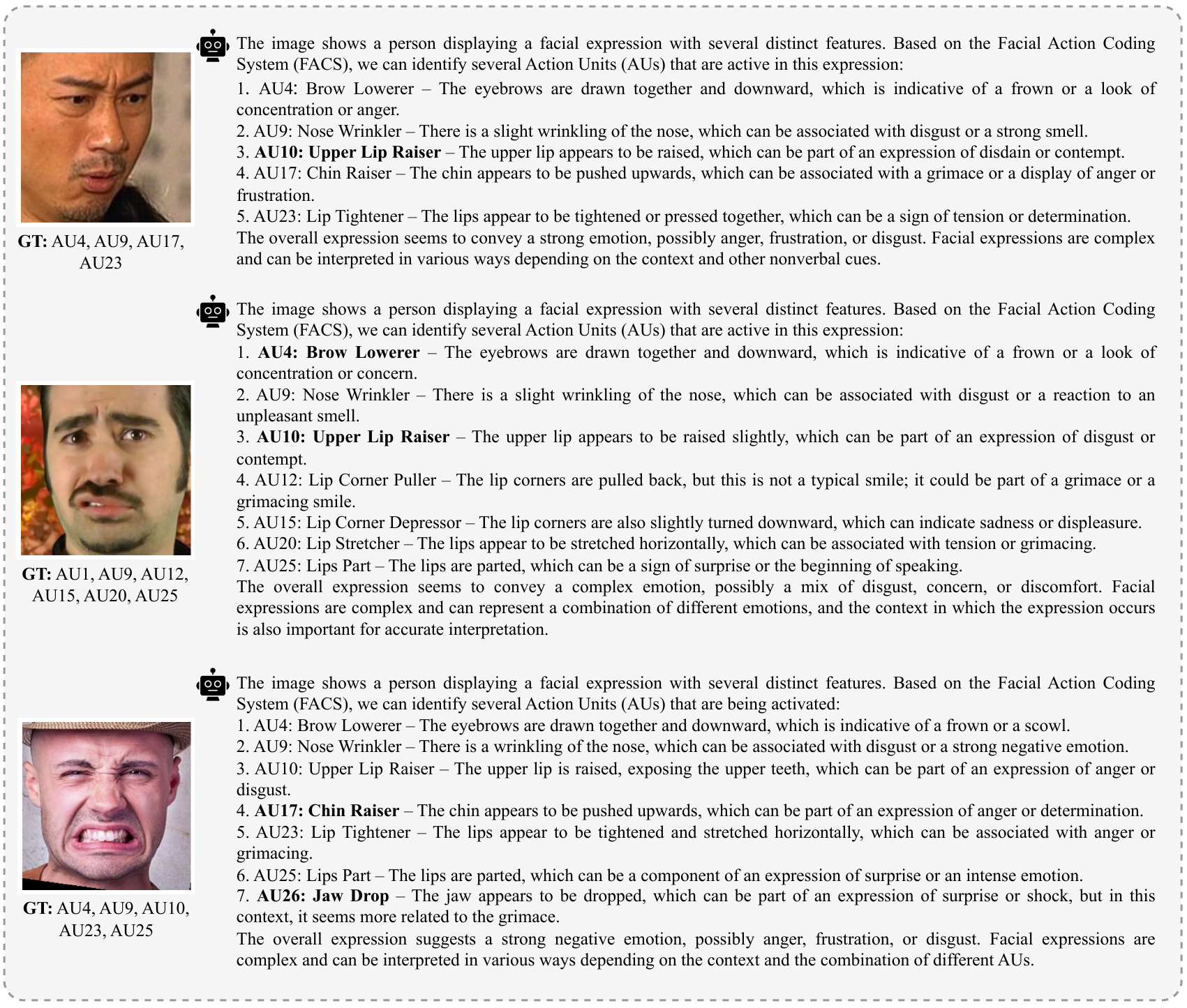}
        \caption{Examples of EmoLA's failed prediction on AU task.}\vspace{10pt}
        \label{fig:au_fail}
    \end{subfigure}
    \caption{Failed cases of EmoLA on FABA-Instruct.}
\end{figure}

We also present some failed cases of EmoLA's prediction on our FABA-Instruct dataset in \cref{fig:emo_fail} and \cref{fig:au_fail}, respectively.

\clearpage

\end{document}